\pdfoutput=1
\documentclass[11pt]{article}

\usepackage[final]{acl}

\usepackage{times}
\usepackage{latexsym}
\usepackage[ruled,vlined]{algorithm2e}

\usepackage{microtype}
\usepackage{graphicx}
\usepackage{subfigure}
\usepackage{booktabs} % for professional tables
\usepackage{makecell}
\usepackage{tikz}
\usepackage{pifont}

\usepackage[ruled,vlined]{algorithm2e}
\usepackage{amsmath}
\usepackage{hyperref}

\usepackage{graphicx}
\usepackage{booktabs}
\usepackage{multirow}
\usepackage{array}
\usepackage{xcolor}

\newcommand{\eat}[1]{}
\newcommand{\eg}{{\em e.g., }}     % e.g.
\newcommand{\ie}{{\em i.e., }}      % i.e.
\newcommand{\cm}{\color{red}}
\newcommand{\cmth}{\color{blue}}

\usepackage[T1]{fontenc}
\usepackage[utf8]{inputenc}

\usepackage{microtype}

\usepackage{inconsolata}

\usepackage{graphicx}

\title{Is Depth All You Need? \\ An Exploration of Iterative Reasoning in LLMs}

\author{
 \textbf{Zongqian Wu\textsuperscript{1,2}},
 \textbf{Tianyu Li\textsuperscript{1}},
 \textbf{Baoduo Xu\textsuperscript{1}},
 \textbf{Jiaying Yang\textsuperscript{1}}, 
  \textbf{Mengmeng Zhan\textsuperscript{1}}, \\
\textbf{Xiaofeng Zhu\textsuperscript{1}}\footnotemark[1],
\textbf{Lei Feng\textsuperscript{2}}\footnotemark[1],
\\
 \textsuperscript{1}UESTC,
 \textsuperscript{2}SUTD,
\\
   {wkzongqianwu@gmail.com} 
}

\begin{document}
\maketitle
\renewcommand{\thefootnote}{*}
\footnotetext[1]{Corresponding authors.}

\begin{abstract}
Deep iterative chain-of-thought (CoT) reasoning enables LLMs to tackle complex tasks by progressively activating relevant pre-trained knowledge. However, it faces challenges in ensuring continual improvement and determining a stopping criterion. 
In this paper, we investigate whether the relevant knowledge that contributes directly to solving the given question can be activated from the initial reasoning path, thus circumventing the need for iterative refinement. Our experiments reveal that increasing the diversity of initial reasoning paths can achieve comparable or superior performance, a concept we term \textit{breadth reasoning}. However, existing breadth reasoning approaches, such as self-consistency, offer limited diversity. To address this limitation, we propose a simple yet effective method that enhances reasoning breadth by integrating contextual exploration with reduced sampling randomness. Extensive experiments demonstrate that our approach significantly outperforms deep iterative reasoning. Our code is provided in \url{https://github.com/zongqianwu/breadth}.
\end{abstract}

\section{Introduction}
\begin{figure}
    \begin{center}
    \includegraphics[width=1\linewidth]{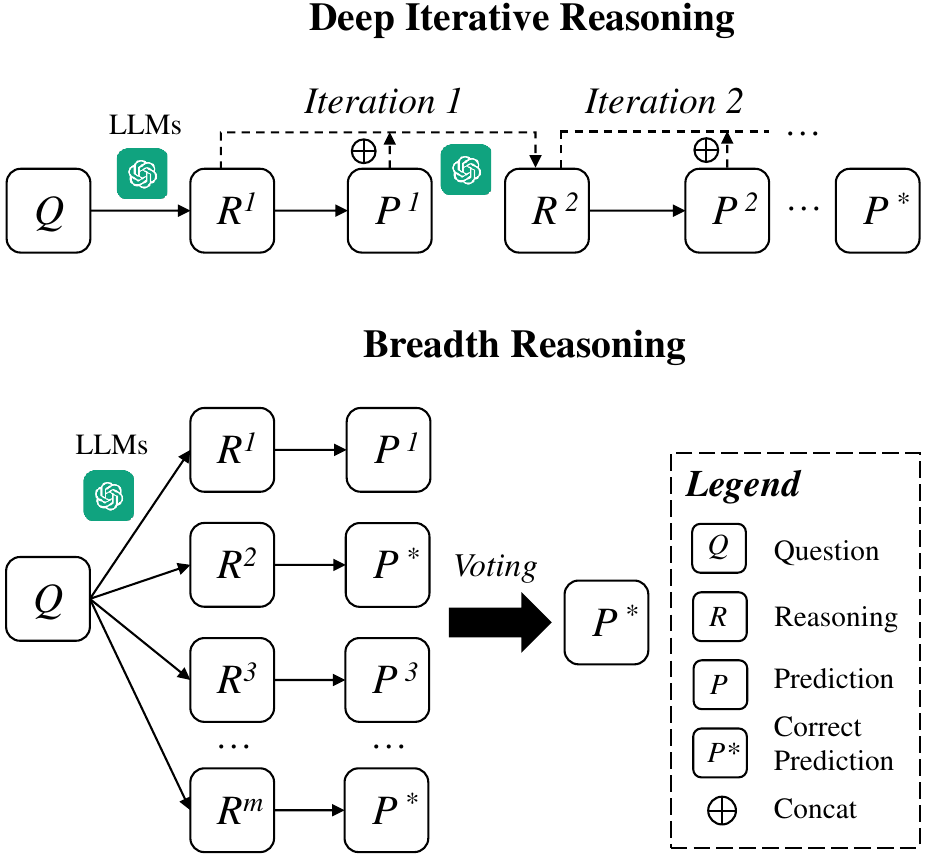}
    \end{center}
    % \vspace{-2pt}
    \caption{
     \textbf{Deep Iterative Reasoning} extends the CoT reasoning by iteratively refeeding both the reasoning steps and predictions into LLMs, while \textbf{Breadth Reasoning} involves generating diverse reasoning paths and aggregating corresponding multiple predictions.
    }
    \label{fig:DR-BR}
    \vspace{-10pt}
\end{figure}

Chain-of-thought (CoT) reasoning improves the performance of large language models (LLMs) on complex tasks by guiding them to construct intermediate steps before producing final answers \cite{wei2022chain, kojima2022large}.
Recently, the OpenAI o1 model \cite{O1} extended CoT by refeeding both the reasoning process and prediction as new inputs into LLMs and constructing multiple iterative rounds. This approach, which further enhances the ability of LLMs to tackle more challenging tasks, is termed \textit{deep iterative reasoning}, as shown in the upper part of Figure \ref{fig:DR-BR}.

However, deep iterative reasoning faces two key challenges: (i) ensuring that the reasoning process in new iteration surpasses the previous one, and (ii) determining the appropriate stopping criterion for the iterations.
To avoid these issues, we first analyze the underlying mechanism of deep iterative reasoning: it encourages LLMs to iteratively utilize model-generated information as self-reminders, aiming to activate relevant pre-trained knowledge that contributes directly to solving the given question.
Based on this observation, we pose a problem: \textit{Can relevant pre-trained knowledge that contributes directly to solving the given question be activated from the initial reasoning path?} 
If so, this would eliminate the need for iterative reasoning, thereby circumventing its associated challenges.

% To address these issues, our experiments observe that the effectiveness of deep iterative reasoning stems from LLMs leveraging model-generated information as self-reminders \cite{wu2024rethinking}, thereby progressively activating pre-trained knowledge relevant to the given test question. 

To investigate the above problem, we conducted a pilot experiment. Specifically, we first selected samples where standard CoT reasoning made incorrect predictions, while deep iterative reasoning made correct predictions. Then, we applied the self-consistency approach \cite{wang2022self}, which generates multiple initial reasoning paths by sampling from LLMs and aggregates the predictions through voting, to re-test these samples using only a single round of iteration. The experimental results yielded two key findings: (i) when the number of sampling paths was set to one, approximately half of the samples were correctly reclassified, and (ii) as the number of sampling paths increased, performance improved initially and then plateaued.
These findings suggest that generating diverse initial reasoning paths 
% and aggregating the corresponding multiple predictions 
has the potential to activate relevant pre-trained knowledge that contributes directly to solving the given question. We refer to this alternative approach as \textit{breadth reasoning}, as illustrated in the lower part of Figure \ref{fig:DR-BR}.

% These findings reveal that when LLMs inherently possess the potential to solve the question, deep iterative reasoning is unnecessary. Instead, increasing the diversity of reasoning paths and aggregating the corresponding predictions can also effectively resolve the question,
% which we refer to as \textit{breadth reasoning}, as shown in the lower part of Figure \ref{fig:DR-BR}.

However, the diversity of reasoning paths generated by the self-consistency approach is limited, failing to fully leverage the potential of LLMs. As indicated by the aforementioned finding (ii), self-consistency alone cannot fully replace the advantages of deep iterative reasoning.
To address this limitation, we reviewed the entire CoT reasoning process and 
identified several potential factors that could contribute to the generation of diverse reasoning paths. First, semantically equivalent but differently expressed modifications of the given question or the pre-defined prompt. Second, perturbations introduced to LLMs during the reasoning generation stage. Finally, multiple sampling of reasoning paths from LLMs (\ie self-consistency).

% These factors can be categorized into three types: (i) a semantically invariant but differently expressed modification of the input question or pre-defined prompt; (ii) perturbations applied to LLMs during the reasoning or prediction generation stages; (iii) changes resulting from the sampling of LLMs during the reasoning or prediction generation stages.

Next, we assess the impact of various factors on the diversity of reasoning paths by using a set of questions that LLMs are unable to solve. 
% This evaluation strategy is designed based on the observation that for simple questions, the different reasoning paths generated by LLMs typically lead to the same prediction. In contrast, for more complex questions, the predictions from different reasoning paths often exhibit greater variability. 
The evaluation show that, aside from the perturbation directly applied to LLMs, the other three factors significantly influence the diversity of reasoning paths. Among these, the diversity induced by self-consistency is smaller compared to the other two factors. 
Building on this insight, we propose a simple yet effective breadth reasoning method to generate diverse reasoning paths.  Specifically, we modify the expression of the given question or prompt while preserving its original meaning, in conjunction with self-consistency. 
Through extensive experiments, we find that this method effectively extend the breadth of reasoning, significantly outperforming deep iterative reasoning.

Our contributions can be summarized as follows:
\begin{itemize}
\item Our experiments reveal that the effectiveness of deep iterative reasoning arises from the progressive activation of relevant pre-trained knowledge through consecutive iterations.

\item We have found that generating diverse initial reasoning paths 
has the potential to activate relevant pre-trained knowledge that contributes directly to solving the given question.

\item We analyzed the factors introducing diverse reasoning paths in the CoT process and proposed a method to extend reasoning breadth, thereby fully leveraging the potential of LLMs and outperforming deep iterative reasoning.
\end{itemize}

% We analyzed the factors that contribute to reasoning path diversity in the CoT process. Based on this analysis, we proposed a trick to broaden reasoning breadth, fully leveraging the potential of LLMs and outperform deep iterative reasoning.

\section{Deep Iterative Reasoning}

% This section provides a detailed analysis of deep iterative reasoning. Specifically, Section \ref{pre} outlines its complete process. Section \ref{PTK} highlights that deep iterative reasoning as a process of pre-trained knowledge. Finally, Section \ref{isdir} investigates whether this knowledge can be activated from the initial reasoning path.

This section provides a detailed analysis of deep iterative reasoning. Specifically, Section \ref{pre} outlines its complete process, while Section \ref{PTK} examines its role in activating prior knowledge. Finally, Section \ref{isdir} investigates whether this knowledge can be directly activated from the initial reasoning path.

% \section{Deep Iterative Reasoning: A Process of Gradually Activating Relevant Pre-trained Knowledge}

\subsection{From Standard CoT to Depth Reasoning}
\label{pre}

Given the $i$-th question $q_i$ of dataset and the prompt $\hat{g}$ (\eg ``\textit{Let's think step by step}''), the CoT reasoning steps can be generated by the LLM:
\begin{equation}
r_{i}^{\prime} = \mathbf{LLM}(\mathbf{Concat}(q_i, \hat{g})), 
\label{eq1}
\end{equation}
where $\mathbf{Concat}(\cdot)$ function refers to the sequential concatenation of the specified texts. Next, the reasoning steps $r_{i}^{\prime}$ obtained from Eq. (\ref{eq1}) is concatenated with the question $q_i$ and the prompt $\hat{g}$. The resulting concatenated texts are then fed into the LLM to generate prediction:
\begin{equation}
p_{i}^{\prime} = \mathbf{LLM}(\mathbf{Concat}(q_i, \hat{g}, r_{i}^{\prime})).
\label{eq2}
\end{equation}
For further deep iterative reasoning mechanism, the current reasoning steps $r_{i}^{\prime}$ and prediction $p_{i}^{\prime}$ are used as feedback and fed back into the LLM, generating a new reasoning steps:
\begin{equation}
r_{i}^{\prime \prime} = \mathbf{LLM}(\mathbf{Concat}(q_i, \hat{g}, r_{i}^{\prime}, p_{i}^{\prime}, g^{	\ast})), 
\label{eq3}
\end{equation}
where $g^{\ast}$ is a prompt used to guide the new reasoning. It can either be a reuse of $\hat{g}$ or be manually constructed \cite{wu2024rethinking}.
Next, the LLM discard the earlier reasoning $r_{i}^{\prime}$ and the prediction $p_{i}^{\prime}$, generating new predictions based solely on $r_{i}^{\prime \prime}$:
\begin{equation}
p_{i}^{\prime \prime} = \mathbf{LLM}(\mathbf{Concat}(q_i, \hat{g}, r_{i}^{\prime \prime})).
\label{eq4}
\end{equation}
The process can be repeated (Eqs. (\ref{eq3}) - (\ref{eq4})) until the pre-defined stopping criteria are met, or the maximum iteration count is reached.

% \subsection{Is Deep Iterative Reasoning Necessary?}

\begin{figure*}
    \begin{center}
    \includegraphics[width=0.99\linewidth]{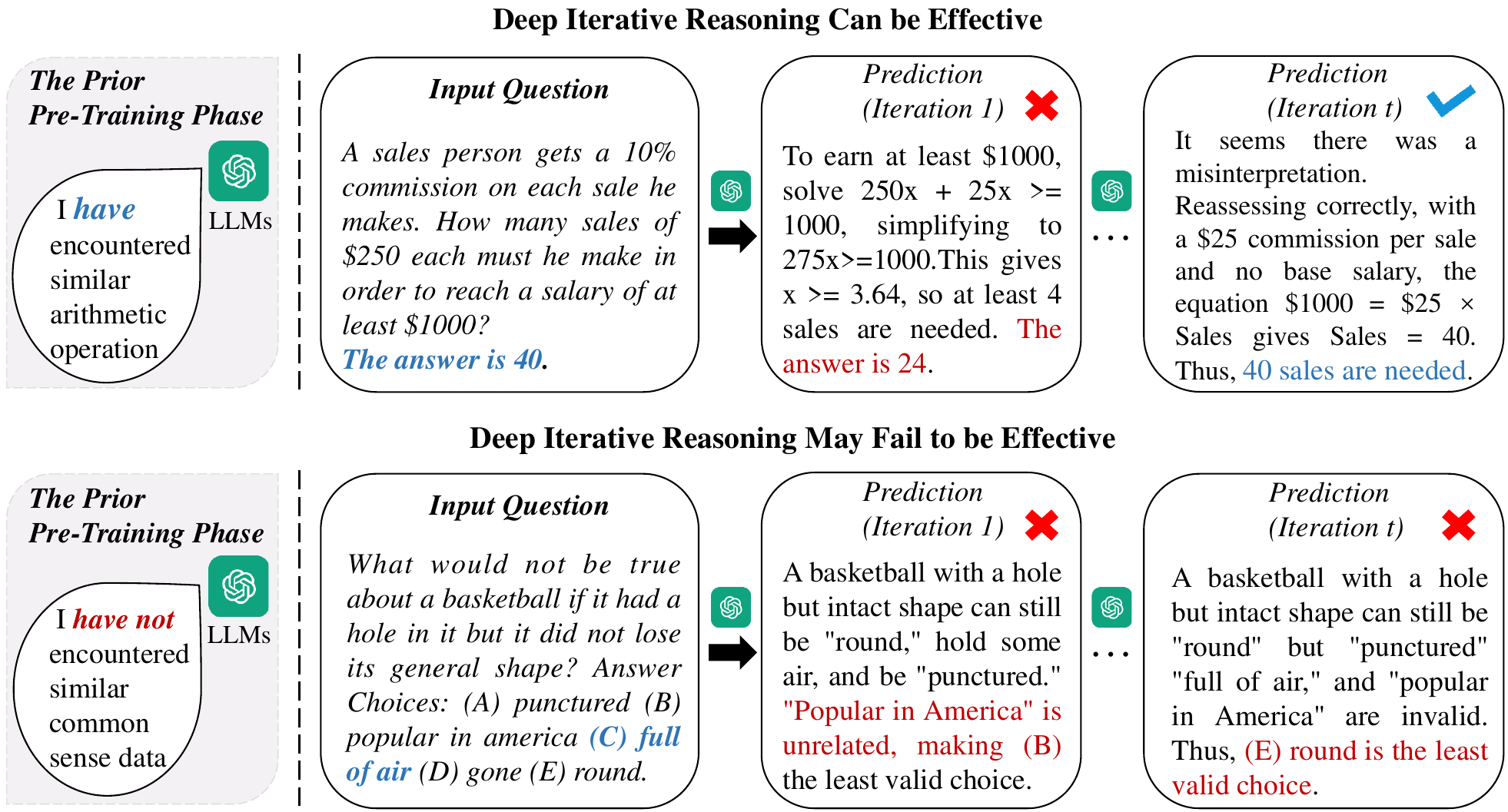}
    \end{center}
    \vspace{-2pt}
    \caption{
    The effectiveness of deep iterative reasoning across different task types. The upper part demonstrates how deep iterative reasoning improves performance on arithmetic tasks, which rely on logical reasoning. As iterations progress, prior logical knowledge is progressively activated. In contrast, the lower part shows limited improvements on commonsense tasks, which primarily depend on information retrieval. If LLMs have not encountered relevant commonsense knowledge during pre-training, deeper reasoning alone is unlikely to resolve these challenges.
    }
    \label{fig:effective}
    \vspace{-8pt}
\end{figure*}

\begin{figure}[htbp]
    \centering
    \subfigure{
        \begin{minipage}[t]{0.47\linewidth} % 减少宽度稍微增加紧凑性
            \centering
            \includegraphics[width=1.02\linewidth]{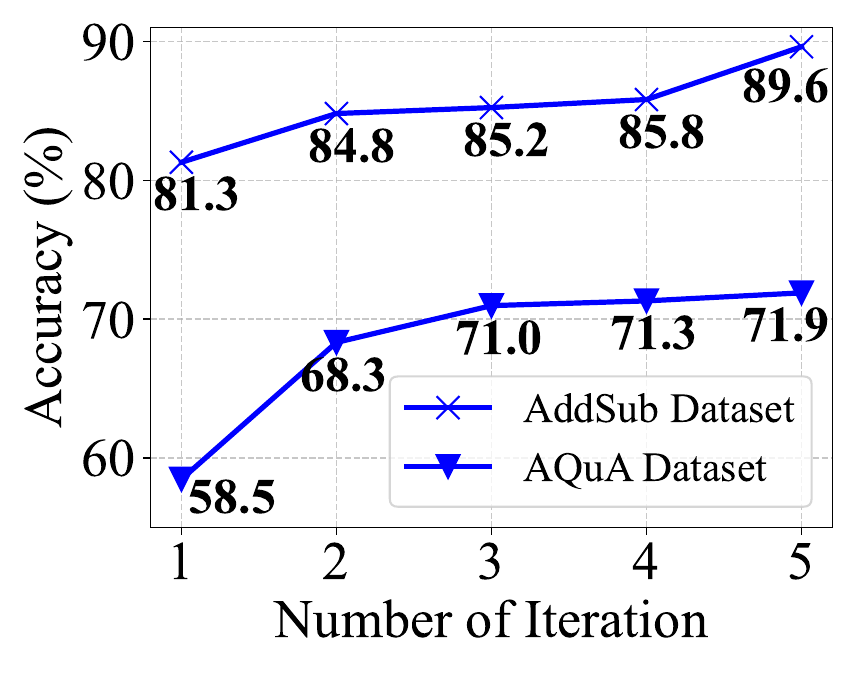}
        \end{minipage}
    } \hspace{-0.5em} % 减少子图间的间距
    \subfigure{
        \begin{minipage}[t]{0.48\linewidth}
            \centering
            \includegraphics[width=1.02\linewidth]{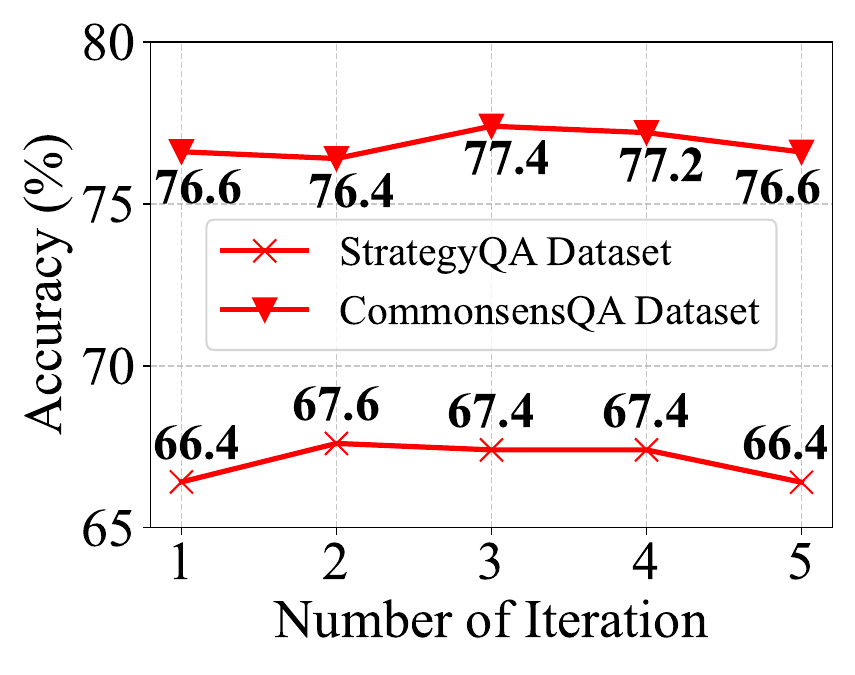}
        \end{minipage}
    }
    \vspace{-10pt} % 减少顶部和子图之间的间距
    \caption{Performance of the LLMs with the deep iterative reasoning mechanisms across varying iterations on \textbf{arithmetic tasks (left)} using the AQuA and AddSub datasets, and \textbf{commonsense tasks (right)} using the StrategyQA and CommonsenseQA datasets.}
    \vspace{-10pt} % 减少 caption 和下方内容之间的间距
    \label{fig:differ_iteration}
\end{figure}

\subsection{Deep Iterative Reasoning as a Gradual Activation of Prior Knowledge}
\label{PTK}

Although deep iterative reasoning, as described in Section \ref{pre}, has demonstrated strong performance \cite{O1, wu2024rethinking}, its effectiveness is not universally applicable across all tasks.

% \footnote{\textbf{Related Work} is provided in the Appendix.}.

As shown in Figure \ref{fig:differ_iteration}, the performance of LLMs on arithmetic tasks (\ie AQuA and AddSub datasets) improves significantly from 69.9\% to 80.8\% as the number of iterations increases. In contrast, the average performance on commonsense tasks (\ie StrategyQA and CommonsenseQA datasets) remains stable, showing no noticeable improvement.
% 改成具体数值分析
The discrepancy arises from  the fundamental nature of these tasks.
Arithmetic questions inherently require logical reasoning, where each reasoning steps and its prediction serve as the basis for the subsequent reasoning, forming a progressive deduction process. As the iterations advance, prior logical knowledge related to question-solving is gradually activated, enabling LLMs to derive increasingly reliable answers.
An example of this is shown in the upper part of Figure \ref{fig:effective}.

However, the lack of performance gains in commonsense tasks can be attributed to 
their core dependence on information retrieval rather than iterative reasoning. For instance, the concept of ``normal basketball'' can be linked to facts such as ``popular in America'' and ``full of air''. If LLMs have not encountered similar commonsense knowledge during the pre-training phase, deeper iterative reasoning alone is unlikely to resolve these challenges. The lower part of Figure \ref{fig:effective} provides an illustrative example of this limitation.

Building on these experimental findings, we draw the following important conclusion: \textit{Deep iterative reasoning does not generate new knowledge beyond what is embedded in pre-trained LLMs. Instead, its effectiveness stems from LLMs leveraging model-generated information as self-reminders, progressively activating existing pre-trained knowledge relevant to the given test question.}

% the gradual activation and reuse of existing prior knowledge through consecutive iterations.

\subsection{Can Depth Reasoning be Replaced?}
\label{isdir}

Based on the conclusion in Section \ref{PTK}, we pose a problem: \textit{Can relevant pre-trained knowledge that directly contributes to solving the given question be activated from the initial reasoning path?}
If so, this would eliminate the need for iterative reasoning, thereby bypassing its associated challenges, \eg progressive refinement and stopping criterion.

\begin{figure}[htbp]
    \centering
    \subfigure{
        \begin{minipage}[t]{0.46\linewidth} % 减少宽度稍微增加紧凑性
            \centering
            \includegraphics[width=1.03\linewidth]{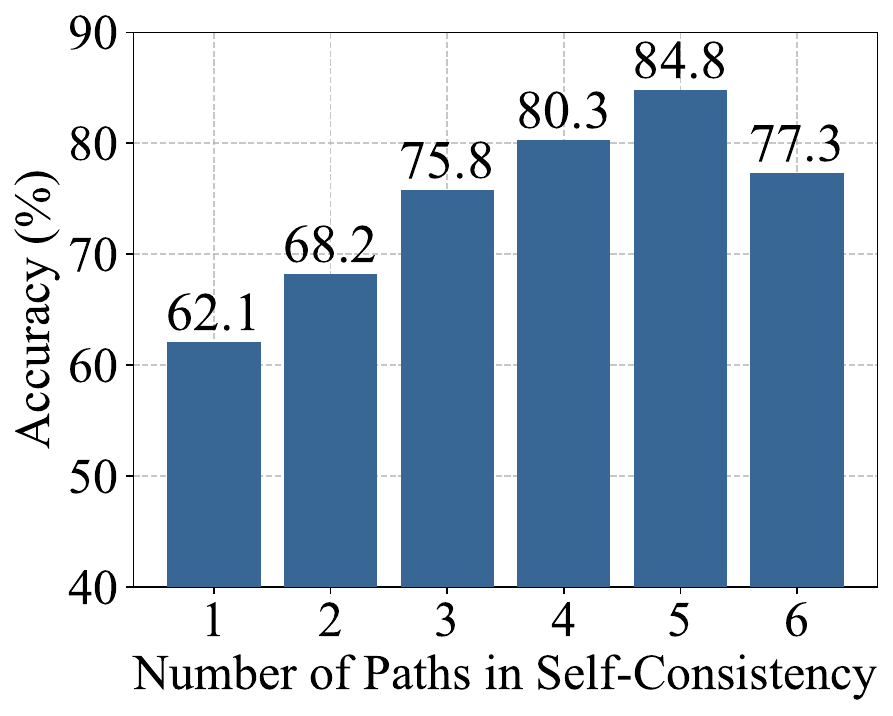}
        \end{minipage}
    } \hspace{-0.6em} % 减少子图间的间距
    \subfigure{
        \begin{minipage}[t]{0.46\linewidth}
            \centering
            \includegraphics[width=1.03\linewidth]{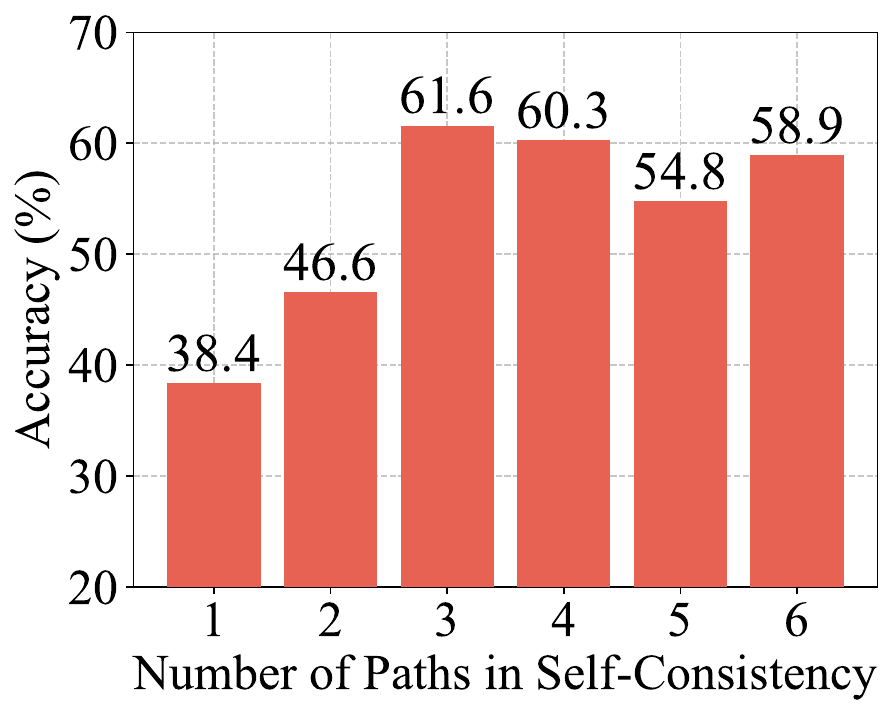}
        \end{minipage}
    }
    % \vspace{-6pt} % 减少顶部和子图之间的间距
    \caption{Performance of self-consistency on a subset of AddSub (left) and AQuA (right) under varying numbers of reasoning paths. This subset consists of samples that were originally misclassified by standard CoT but were correctly predicted through deep iterative reasoning.}
    \vspace{-4pt} % 减少 caption 和下方内容之间的间距
    \label{fig:incorrect}
\end{figure}

To investigate this issue, we first selected samples from the arithmetic experiment (\ie AQuA and AddSub datasets) constructed in Section \ref{PTK}, where standard CoT reasoning produced incorrect predictions, while deep iterative reasoning made correct predictions. 
Subsequently, we applied the self-consistency approach, which generates diverse reasoning paths by sampling from LLMs and aggregates the corresponding multiple predictions through majority voting, to re-test these samples using only a single round of iteration. The experimental results are shown in Figure \ref{fig:incorrect}.
For the AddSub dataset, accuracy exhibits a steady upward trend with the increase in the number of reasoning paths, reaching a peak at five paths (84.8\%), after which it slightly decreases at six paths (77.3\%). Similarly, in the AQuA dataset, accuracy improves from 38.4\% with a single reasoning path to 61.6\% at three paths. However, beyond three paths, the improvement plateaus, with accuracy stabilizing between four and six paths, peaking at 58.9\%.

These experimental results can be distilled into two key findings: (i) with only one reasoning path, approximately half of the samples are correctly reclassified; and (ii) as the number of reasoning paths increases, the performance initially improves and then stabilizes.
This indicates that generating diverse initial reasoning paths has the potential to activate relevant pre-trained knowledge that contributes directly to solving the given question.
Subsequently, the correct answer can be determined through major voting across predictions from different reasoning paths.
The underlying rationale is that a given question may have multiple solutions.
% , and if LLMs have the potential to solve it, most solutions will typically converge to the same prediction. 
% In such cases, the correct prediction can be effectively selected through a majority voting strategy. 
We refer this alternative approach as \textit{breadth reasoning}, in contrast to deep iterative reasoning.

However, the diversity of reasoning paths generated through self-consistency is limited, restricting the breadth of reasoning and thus failing to fully leverage the potential of LLMs.
As shown in Figure \ref{fig:incorrect}, 
self-consistency alone does not completely replace the advantages of deep iterative reasoning.
We will address this limitation in Section \ref{br}.

% To address this limitation, we will review the entire CoT reasoning process in Section \ref{br} and identify factors that may contribute to the generation of diverse reasoning paths. Based on these factors, we propose a simple trick to expand the breadth of reasoning, thereby fully leveraging the potential of LLMs.

\section{Breadth Reasoning}
\label{br}

In this section, we begin by reviewing the entire CoT reasoning process, identifying factors that may contribute to the generation of diverse reasoning paths, as discussed in Section \ref{influ}. Then, we investigate the impact of these factors in Section \ref{aif}. Finally, Section \ref{trick} introduces a simple yet effective method to expand the reasoning breadth.

\subsection{What Influences Reasoning Diversity?}

\label{influ}

\begin{figure*}
    \begin{center}
    \includegraphics[width=0.96\linewidth]{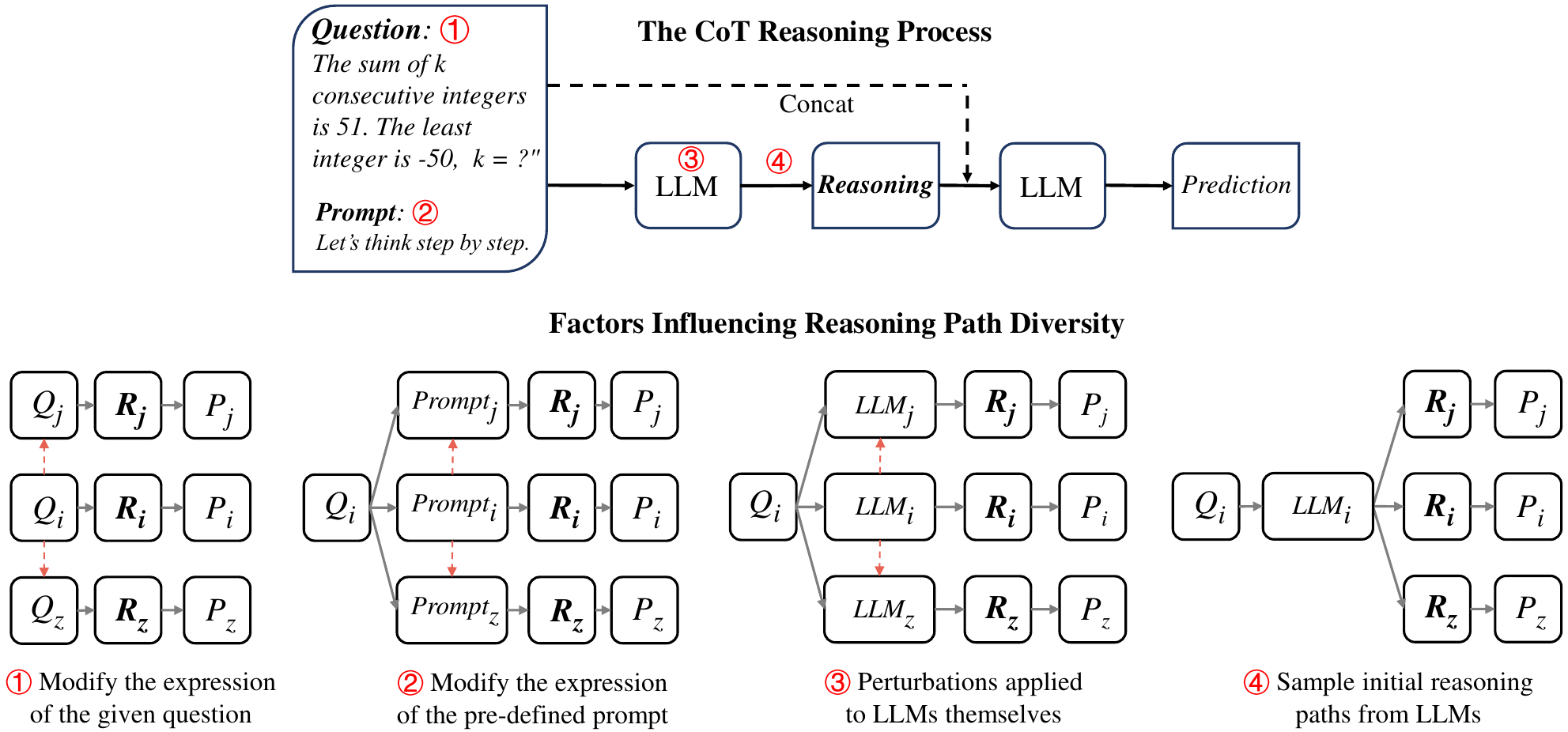}
    \end{center}
    \vspace{-2pt}
    \caption{
    In the upper part, we review the complete CoT reasoning process, where the given question is concatenated with the pre-defined prompt and then fed into LLMs for reasoning, ultimately producing the corresponding prediction. During this process, four factors that may contribute to diverse initial reasoning paths are identified as follows:  
    Modify the expression of the given question {\cm \ding{192}} or the pre-defined prompt {\cm \ding{193}}; Perturbations applied to LLMs themselves {\cm \ding{194}}; and Sample initial reasoning paths from LLMs {\cm \ding{195}}. The lower part further elaborates on how these factors contribute to the diversification of initial reasoning paths, leading to multiple predictions.
    }
    \label{fig:differ_fac}
    \vspace{-6pt}
\end{figure*}

We review the complete CoT reasoning process in the upper part of Figure \ref{fig:differ_fac}. As the CoT process progresses, we identify four critical factors that may generate diverse initial reasoning paths.

Specifically, the CoT process begins with the given question and pre-defined prompt. Modifying the expression of these elements, while preserving their original meaning, can lead to different reasoning paths, as illustrated in \ding{192} and \ding{193} of Figure \ref{fig:differ_fac}. Next, the fixed question and prompt are input into the LLMs, where perturbations to the LLMs themselves 
% such as parameter perturbations (\eg adjusting weights) or embedding perturbations (\eg adding random noise to sentence embeddings), 
can also influence the resulting reasoning paths, as shown in \ding{194} of Figure \ref{fig:differ_fac}. Finally, LLMs generate reasoning paths based on the question and prompt. By performing multiple samplings, different reasoning paths can be explored, similar to self-consistency, as depicted in \ding{195} of Figure \ref{fig:differ_fac}.

% We then evaluate the impact of these factors on the diversity of reasoning paths by using a set of questions that LLMs are unable to solve. 
% This evaluation strategy is designed based on the observation that for simple questions, the different reasoning paths generated by LLMs typically lead to the same prediction. In contrast, for more complex questions, the predictions from different reasoning paths often exhibit greater variability. 
% The experimental results are presented in Table 1.

\subsection{Impact Analysis of Factors}

\label{aif}

We assess the impact of these factors on the diversity of reasoning paths using a set of questions that LLMs cannot solve. This evaluation involves generating multiple predictions through diverse reasoning paths shaped by these factors. To quantify diversity, we compute the entropy of these predictions, where a higher entropy value signifies that the factor fosters a broader exploration of reasoning paths. This analysis strategy is grounded in the observation that, for simple questions, the different reasoning paths generated by LLMs typically lead to the same prediction. However, for more complex questions, the predictions derived from different reasoning paths tend to exhibit greater variability. 
% The results are presented in Table 1 in Appendix.

The evaluation results shown in Table \ref{tab:entropy} in Appendix indicate that, apart from the perturbation directly applied to LLMs, the other three factors significantly influence the reasoning path diversity. Among them, the diversity induced by self-consistency is smaller compared to the other two factors. 
The fundamental reason for this difference lies in 
% varying the expression of the question or prompt encourages LLMs to explore different hypotheses, consider diverse premises, or initiate reasoning from different sequences. Additionally, since 
LLM's strong dependence on context. Effectively modifying the expression of a question or prompt is equivalent to creating a new context, prompting the model to reassess the question and generate reasoning paths distinct from the original expression. In contrast, merely sampling reasoning paths from LLMs remains within the same contextual framework, leading to limited variation and failing to overcome cognitive inertia effectively.

Based on the above analysis, we draw two important insights: (i) perturbing LLMs themselves has little impact on the diversity of initial reasoning path, as LLMs possess strong robustness; (ii) the diversity of reasoning paths guided by factors later in the CoT reasoning process decreases, as the input context becomes progressively fixed, reducing the degrees of freedom available to LLMs.

% Specifically, for the input question and a pre-defined prompt, we additionally use LLMs to generate new expressions, thereby producing multiple predictions, as shown in \ding{192} and \ding{193} of Figure \ref{fig:differ_fac}. For perturbations within LLMs, we perturb the internal sentence embeddings of LLMs to obtain multiple predictions, as shown in \ding{194} of Figure \ref{fig:differ_fac}. For the generation of reasoning paths, we fix the temperature coefficient at 0.8 and perform multiple rounds of sampling to generate corresponding predictions, as shown in \ding{195} of Figure \ref{fig:differ_fac}. Finally, we calculate the information entropy of the predictions corresponding to the four factors. A higher entropy value indicates that the factor leads to more diverse reasoning paths.

\subsection{Proposed Method}

\label{trick}

Building on the insights in Section \ref{aif}, we propose a sample yet effective method to extend the reasoning breadth. Specifically, we modify the expression of the given question or pre-defined prompt while preserving its original meaning.
These input context modifications encourages LLMs to explore different hypotheses, consider diverse premises, or initiate reasoning from different sequences.
% Each reformulated expression serves as the foundation for an independent CoT reasoning process.

To achieve this, we first construct an instruction:
\textbf{\# Instruction:} \textit{Rephrase the following sentence to change its wording and structure while maintaining the same meaning. Ensure the core sentence remains unchanged.
}
Next, we concatenate this instruction with the question or prompt and feed it into LLMs to generate multiple reformulated versions.
Each reformulation serves as the foundation for an independent CoT reasoning process.

% These input context modifications encourages LLMs to explore different hypotheses, consider diverse premises, or initiate reasoning from different sequences. 

During each CoT process, we generate multiple reasoning paths, thereby reducing sampling randomness.
By integrating contextual exploration with self-consistency, we extend the reasoning breadth and fully leverage the potential of LLMs.

% ensuring more stable outputs when the context remains fixed. 
% By integrating context exploration with reduced sampling randomness, we effectively extend the breadth of reasoning.

For instance, if we generate three distinct reformulations of a given question and apply self-consistency sampling twice per reformulation, we obtain a total of 
``3×2=6'' diverse reasoning paths.
The final answer is then selected using voting based on the six predictions from these reasoning paths.

% By integrating context exploration with reduced sampling randomness, we broaden the breadth of reasoning and fully leverage the potential of LLMs.

\begin{table*}[htbp]
  \centering
  \resizebox{\textwidth}{!}{
  \begin{tabular}{lccccccccccc}
    \toprule
    \textbf{Method} & \multicolumn{6}{c}{\textbf{Arithmetic}} & \multicolumn{2}{c}{\textbf{Common}} & \multicolumn{2}{c}{\textbf{Symbolic}} & \textbf{Overall} \\
    \cmidrule(lr){2-7} \cmidrule(lr){8-9} \cmidrule(lr){10-11}
    & Multi & GSM8K & Single & Add & AQuA & SVAM & ST & CS & Letter & Coin & Avg. \\
    \midrule
    Zero-Shot & 51.2 & 10.8 & 62.4 & 56.7 & 36.6 & 56.3 & 66.2 & 74.5 & 1.4 & 50.2 & 46.6 \\
    CoT & 92.8 & 74.7 & 84.4 & 74.7 & 55.5 & 77.0 & 63.5 & 73.6 & 55.0 & 93.4 & 74.5 \\
    \hline
    Deep-CoT & 94.7 & 81.6 & 90.6 & 84.2 & 66.5 & 83.5 & 67.5 & 75.6 & 68.9 & 96.0 & 80.9 \\
    ARI & 96.7 & 82.6 & 92.1 & \textbf{87.1} & 69.3 & 87.1 & 67.5 & 77.5 & 75.8 & 97.2 & 83.3 \\
    \hline
    SC & 95.7 & 79.2 & 88.8 & 81.3 & 63.0 & 82.2 & 65.9 & 75.3 & 66.2 & 97.2 & 79.5 \\
    QuestionC & 95.3 & 80.8 & 88.8 & 82.5 & 63.4 & 82.9 & 64.1 & 75.8 & 66.0 & 97.8 & 79.7 \\
    PromptC & 94.5 & 81.6 & 92.3 & 85.8 & 64.2 & 83.0 & 66.2 & 76.7 & 67.0 & 96.6 & 80.8 \\
    \hline
    QuestionC-SC & 95.7 & 83.2 & 91.7 & 85.3 & 70.9 & 87.2 & 66.4 & 77.0 & 76.2 & 96.8 & 83.0 \\
    & \textcolor{blue!90}{(+0.0)} & \textcolor{blue!90}{(+4.0)} & \textcolor{blue!90}{(+2.9)} & \textcolor{blue!90}{(+4.0)} & \textcolor{blue!90}{(+7.9)} & \textcolor{blue!90}{(+5.0)} & \textcolor{blue!90}{(+0.5)} & \textcolor{blue!90}{(+1.7)} & \textcolor{blue!90}{(+10.0)} & \textcolor{blue!90}{(-0.4)} & \textcolor{blue!90}{(+3.6)} \\
& \textcolor{brown}{(-1.0)} & \textcolor{brown}{(+0.6)} & \textcolor{brown}{(-0.4)} & \textcolor{brown}{(-1.8)} & \textcolor{brown}{(+1.6)} & \textcolor{brown}{(+0.1)} & \textcolor{brown}{(-1.1)} & \textcolor{brown}{(-0.5)} & \textcolor{brown}{(+0.4)} & \textcolor{brown}{(-0.4)} & \textcolor{brown}{(-0.2)} \\
    PromptC-SC & \textbf{96.7} & \textbf{84.9} & \textbf{93.5} & 86.6 & \textbf{76.4} & \textbf{87.7} & \textbf{68.9} & \textbf{78.4} & \textbf{79.4} & \textbf{98.6} & \textbf{85.1} \\
& \textcolor{blue!90}{(+1.0)} & \textcolor{blue!90}{(+5.7)} & \textcolor{blue!90}{(+4.7)} & \textcolor{blue!90}{(+5.3)} & \textcolor{blue!90}{(+13.4)} & \textcolor{blue!90}{(+5.5)} & \textcolor{blue!90}{(+3.0)} & \textcolor{blue!90}{(+3.1)} & \textcolor{blue!90}{(+13.2)} & \textcolor{blue!90}{(+1.4)} & \textcolor{blue!90}{(+5.6)} \\
& \textcolor{brown}{(+0.0)} & \textcolor{brown}{(+2.3)} & \textcolor{brown}{(+1.4)} & \textcolor{brown}{(-0.5)} & \textcolor{brown}{(+7.1)} & \textcolor{brown}{(+0.6)} & \textcolor{brown}{(+1.4)} & \textcolor{brown}{(+0.9)} & \textcolor{brown}{(+3.6)} & \textcolor{brown}{(+1.4)} & \textcolor{brown}{(+1.8)} \\
    \bottomrule
  \end{tabular}}
  % \vspace{-8pt}
  \caption{Accuracy (\%) on ten reasoning datasets from three categories of reasoning tasks. The maximum number of iterations for the deep iterative reasoning methods and the number of reasoning paths for the breadth reasoning methods are both set to 3. \textcolor{blue!90}{Blue} and \textcolor{brown}{Brown} fonts represent comparisons between our proposed method and the classic breadth reasoning method (\ie self-consistency (SC)), as well as the best-performing deep iterative reasoning method (\ie adaptive reasoning iteration (ARI)). Bold font highlights the best performance in each column.}
  \label{tab:addlabel}
  \vspace{-8pt}
\end{table*}

\section{Experiments}

\subsection{Experimental Settings}
We evaluate our method on ten reasoning datasets, including six arithmetic datasets (\ie MultiArith \cite{roy2016solving}, GSM8K \cite{cobbe2021training}, SingleEq \cite{koncel2015parsing}, AddSub \cite{hosseini2014learning}, AQuA \cite{ling2017program}, and SVAMP \cite{patel2021nlp}), two commonsense reasoning datasets (\ie StrategyQA \cite{geva2021did} and CommonsenseQA \cite{talmor2018commonsenseqa}), and two symbolic reasoning datasets (\ie LastLetter and CoinFlip \cite{wei2022chain}).

% We utilize GPT-3.5-turbo-0125 as the foundation model for all experiments.
% Our comparative experiments encompass three categories of methods. The first category consists of conventional CoT methods, where the question is directly fed into LLMs without any prompts for inference (\ie zero-shot). Based on this, a generic prompt is introduced, and answers are generated using greedy decoding (\ie zero-shot CoT \cite{kojima2022large}).

% The second category includes deep iterative reasoning methods. Specifically, the current reasoning process and predictions are serve as new inputs for the next iteration, continuing until a pre-defined maximum number of iterations is reached (\ie deep reasoning). Furthermore, a dynamic stopping strategy based on the semantic entropy of predictions is incorporated. (\ie adaptive reasoning iteration (ARI) \cite{wu2024rethinking}).
% The third category comprises breadth reasoning methods
% , which aim to obtain multiple predictions by increasing the diversity of reasoning and selecting the final answer through voting. 
% First, we perform multiple sampling of reasoning paths from LLMs (\ie self-consistency). Next, we generate paraphrased versions of the input question or prompt that preserve the original semantics while varying the expression. Finally, building upon self-consistency, we further integrate the aforementioned paraphrasing strategy to improve the robustness of reasoning.

For all experiments, we utilize GPT-3.5-turbo-0125 as the foundation model, chosen for its accessibility and cost-effectiveness. Our comparative study examines three categories of methods:

\begin{itemize}
    \item \textbf{Standard CoT methods}: Directly feed the question into LLMs without prompts for zero-shot inference, then introduce a generic prompt for reasoning with greedy decoding (\ie Zero-Shot CoT \cite{kojima2022large}).
    
    \item \textbf{Deep iterative reasoning methods}: Iteratively refine reasoning by using previous steps and predictions as new inputs, terminating when reaching a pre-defined limit or a semantic entropy-based stopping criterion (\ie Deep-CoT and ARI \cite{wu2024rethinking}).
    
    \item \textbf{Breadth reasoning methods}: Improve robustness by generating diverse predictions and selecting the answer via voting. This includes sampling multiple reasoning paths (\ie self-consistency) and modifying the question or prompt (\ie question-consistency and prompt-consistency) while preserving semantics.
\end{itemize}

\subsection{Evaluating Different Breadth Methods}
Rows 5–9 of Table 1 present the performance of various breadth reasoning methods across ten datasets, including self-consistency (SC), question-consistency (QuestionC), and prompt-consistency (PromptC), as well as our proposed methods, QuestionC-SC and PromptC-SC. The results show that SC performs particularly well on symbolic reasoning tasks (\ie LastLetter and CoinFlip), confirming its effectiveness in maintaining consistency across multiple reasoning paths. Meanwhile, QuestionC and PromptC enhance the diversity of reasoning paths by modifying the expressions of the question and prompt, respectively, leading to further improvements in breadth reasoning performance, with notable gains in arithmetic tasks.

Compared to other breadth reasoning methods, our QuestionC-SC and PromptC-SC achieve superior performance. 
These two methods further explore the contextual space while reducing sampling randomness based on SC.
Notably, PromptC-SC attains the highest average accuracy of 85.1\% across all datasets. The 4.3\% improvement of PromptC-SC over PromptC underscores the effectiveness of integrating SC with contextual prompt space exploration in enhancing reasoning diversity.

% SC+InputC improves the adaptive input mechanism, achieving state-of-the-art results in datasets such as GSM8K and SVAMP. Similarly, SC+PromptC effectively integrates prompt adaptation into the breadth reasoning framework, leading to the highest average accuracy (85.1\%) across all datasets. Notably, SC+PromptC surpasses PromptC by an average of 1.8\%, highlighting the synergy between self-consistency and prompt customization.

Furthermore, in both arithmetic and symbolic tasks (\eg AQuA and LastLetter), QuestionC-SC and PromptC-SC significantly outperform their individual counterparts. This demonstrates the adaptability of our hybrid approaches in handling diverse reasoning challenges. Overall, the results suggest that incorporating self-consistency with contextual space exploration is a promising direction for improving breadth reasoning performance.

\subsection{Breadth v.s. Depth}

We compare breadth reasoning methods with deep iterative reasoning methods (\ie Deep-CoT and ARI) to analyze the respective advantages of these two reasoning paradigms. Experimental results indicate that Deep-CoT performs particularly well on arithmetic tasks such as GSM8K and AQuA datasets, demonstrating its capability to optimize the reasoning process through multiple iterations. Building on this, ARI further introduces an adaptive iterative reasoning mechanism, achieving the best performance (87.1\%) on the AddSub dataset.

However, in symbolic and commonsense reasoning tasks, breadth reasoning methods generally outperform deep iterative reasoning methods. For instance, PromptC-SC surpasses ARI by 3.6\% and 1.4\% on the LastLetter and CoinFlip datasets, respectively. Moreover, PromptC-SC also achieves a higher overall average accuracy (85.1\%) compared to ARI (83.3\%), with an improvement of 1.8\%. These results suggest that breadth reasoning exhibits significant advantages in tasks requiring diverse and structured reasoning paths.

Notably, QuestionC-SC and PromptC-SC also demonstrate strong competitiveness in tasks traditionally favoring deep iterative reasoning methods, such as GSM8K and AQuA datasets. This observation indicates that while deep reasoning excels in logical tasks, breadth reasoning methods can achieve comparable or even superior performance by integrating self-consistency with contextual space exploration. These findings further highlight the significant potential of breadth reasoning methods in handling diverse reasoning tasks.

\subsection{Breadth Analysis}

While our method has shown significant improvements, several issues remain to be explored:
(i) \textit{Does our method truly extend the reasoning breadth?} and (ii)
\textit{Is our method broader than self-consistency, or does it simply gain improvements by increasing the number of reasoning paths?}

% \begin{itemize}
% \item \textbf{\textit{Does our proposed method truly extend the breadth of reasoning path?}}

% \item \textbf{\textit{Is our method broader than self-consistency, or does it simply gain improvements by increasing the number of reasoning paths?}}
% % \item \textbf{\textit{How broad must our method be to surpass deep iterative reasoning?}}
% \end{itemize}

\begin{figure}[htbp]
    \centering
    \subfigure{
        \begin{minipage}[t]{0.475\linewidth} % 减少宽度稍微增加紧凑性
            \centering
            \includegraphics[width=1.02\linewidth]{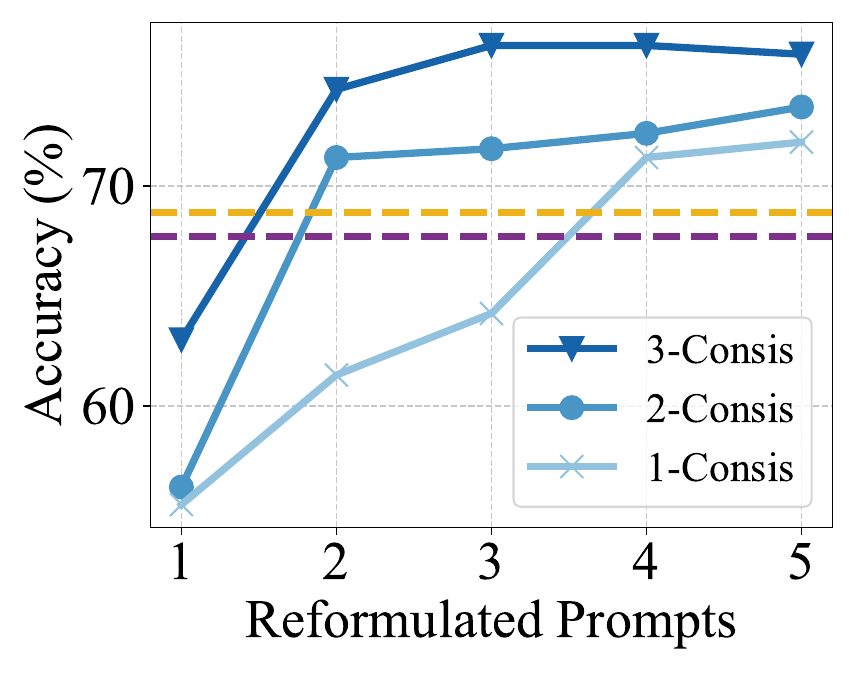}
        \end{minipage}
    } \hspace{-0.5em} % 减少子图间的间距
    \subfigure{
        \begin{minipage}[t]{0.475\linewidth}
            \centering
            \includegraphics[width=1.02\linewidth]{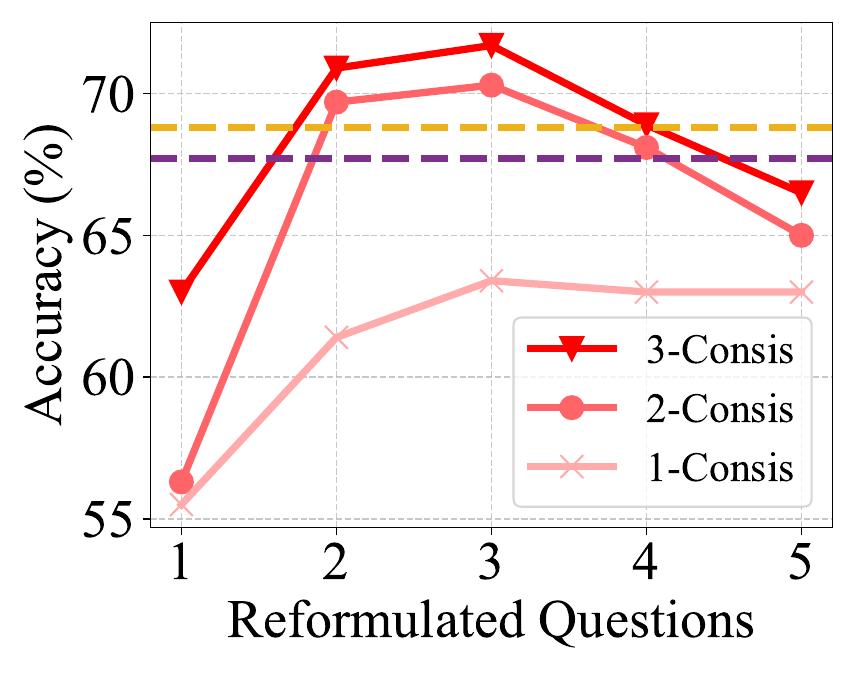}
        \end{minipage}
    }
    \vspace{-16pt} % 减少顶部和子图之间的间距
    \caption{Performance of our breadth reasoning method on the AQuA dataset, evaluated with varying numbers of reformulated prompts and questions, combined with different levels of self-consistency sampling.}
    \vspace{-12pt} % 减少 caption 和下方内容之间的间距
    \label{fig:breadth_analysis}
\end{figure}

To investigate these issues, we conduct two targeted experiments on the AQuA dataset: (i) we vary the number of the given question and prompt reformulations, as well as the number of reasoning path samples, to analyze their impact on performance; (ii) we compare our approach against a self-consistency baseline where the number of reasoning paths is controlled to match that of our method.
The results are illustrated in Figure \ref{fig:breadth_analysis}.

The experimental findings reveal a clear performance improvement as the number of reformulations increases. Notably, when prompt reformulations reaches five, the performance gains begin to plateau. In contrast, when question reformulations reaches four, performance starts to decline. This suggests that modifying the prompt is generally more effective than reformulating the question in exploring the contextual space, as it encourages the LLMs to engage in reasoning from different perspectives. These results address the first research question, confirming that increasing the number of context reformulations enables our method to effectively expand the reasoning paths.

Another key observation is derived from the self-consistency experiments. When the number of reasoning paths in self-consistency reaches six (yellow dashed line) and nine (purple dashed line), the corresponding performance levels are 68.1\% and 67.1\%, respectively. In contrast, our method surpasses these performance levels with only two reformulations combined with two samples (yielding a total of four reasoning paths). This finding suggests that self-consistency, despite increasing the number of reasoning paths, provides only a limited expansion of reasoning diversity.  In contrast, our approach significantly enhances the coverage of reasoning space. These results address the second research question, demonstrating that our method provides a more substantial improvement in reasoning breadth compared to self-consistency.

% For instance, in our approach, we generate a total of ``3×3=9'' reasoning paths through three question reformulations and three path samples per reformulation. Therefore, in the self-consistency setting, we directly set the number of reasoning path samples to 9 and evaluate its performance separately. The experimental results are presented in Figure 1.

\section{Why Our Method Works?}

Our proposed method significantly outperforms existing breadth reasoning and deep iterative reasoning approaches. In this section, we analyze the underlying reasons for these improvements.

\begin{figure}
    \begin{center}
    \includegraphics[width=0.85\linewidth]{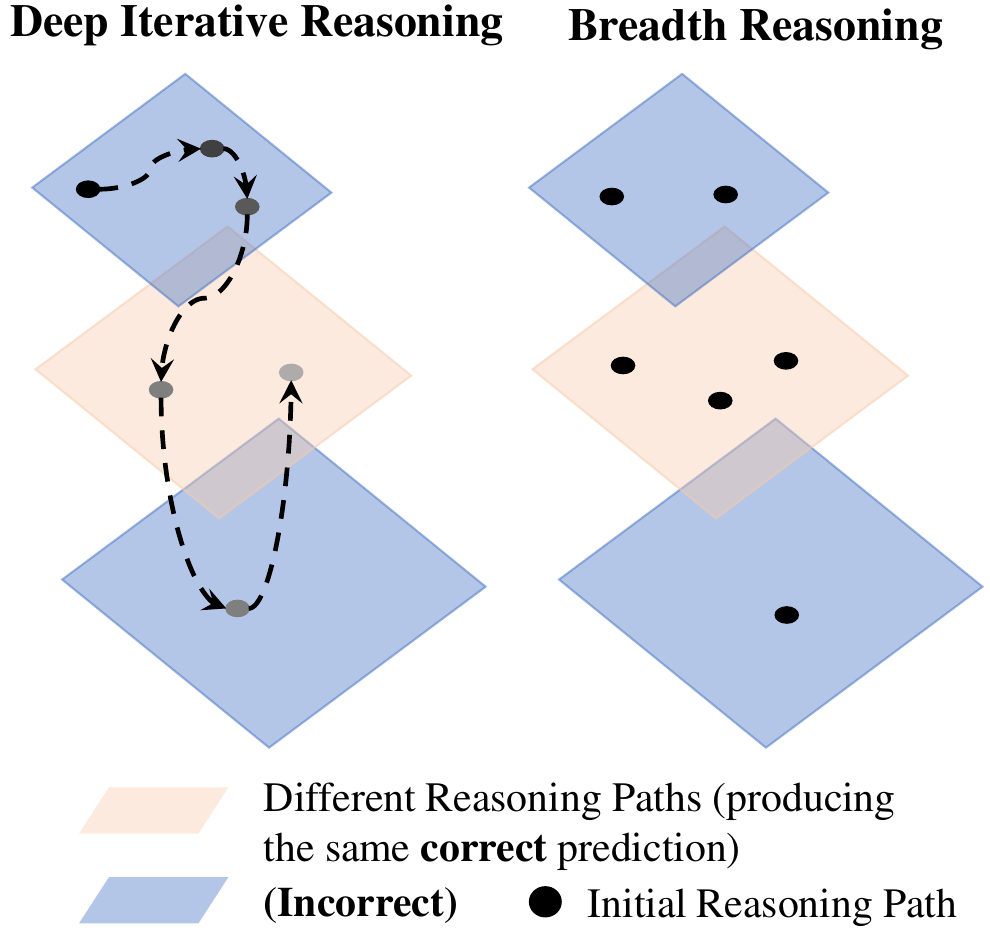}
    \end{center}
    % \vspace{-2pt}
    \caption{
    Visualization of \textbf{Deep Iterative Reasoning} and \textbf{Breadth Reasoning} in identifying the correct reasoning path. Each point on the plane represents an individual reasoning path, all of which eventually converge to the same prediction. 
    In deep iterative reasoning, if the initial reasoning path resides on an incorrect reasoning plane, the process is progressively refined through multiple iterations, eventually transitioning to the correct plane. In contrast, breadth reasoning simultaneously generates multiple initial reasoning paths and determines the final prediction through majority voting.}
    \label{fig:why}
    \vspace{-6pt}
\end{figure}

In breadth reasoning, the diversity of reasoning paths obtained solely through self-consistency (SC) is limited. This reason is that the input context space is fixed, constraining the degrees of freedom in LLMs.  
As illustrated on the right side of Figure \ref{fig:why}, reasoning paths obtained through SC sampling may be confined to the same plane, where different points on the plane represent distinct reasoning paths leading to the same predicted outcome. 

For simple questions, SC-guided reasoning paths typically cluster within the correct plane, meaning that even a single sampled reasoning path is often sufficient to reach the correct answer. In such cases, while SC is effective, it is not essential. However, for complex questions, if the fixed context primarily guides reasoning paths toward an incorrect plane, the SC approach fails.
In contrast, our method enhances the diversity of reasoning paths by exploring the context space, thereby covering a broader range of reasoning strategies. Meanwhile, we integrate SC to improve the reasoning stability under the same context. As a result, our approach increases the probability that the reasoning paths fall within the correct reasoning plane.

On the other hand, deep iterative reasoning progressively refines reasoning process through multiple iterations to achieve the correct prediction. However, it face two critical challenges: (1) how can we ensure that each newly generated reasoning path improves upon the previous iteration? (2) how can we effectively determine the optimal stopping criterion for iteration? As illustrated on the left side of Figure \ref{fig:why}, when a newly generated reasoning path fails to introduce meaningful updates, the iteration becomes ineffective, potentially preventing the discovery of the correct reasoning path within the pre-defined maximum number of iterations. Furthermore, even if the reasoning path has already reached the correct plane, continued iterations may introduce unnecessary extensions, increasing the risk of errors in the final prediction. 

In contrast, our method employs a single round of reasoning strategy, effectively circumventing the limitations of iterative refinement and thereby substantially improving reasoning performance.

% 从那三个角度展开分析，比如一开始深度推理初始化到了一个错误的角度，随着迭代推进，错误不断被纠正。最终问题被正确解决。

% 而宽度推理增加推理宽度可以一开始就很大概率去初始化有效的角度，因此直接解决问题。  给个图例子

% 2. 仅有self-consistency时多样性有限，容易挤在同一个平面内。提升上下文空间后容易在正确的平面产生正确答案，并且保持SC能减少随机性

\section{Conclusion and Future Work}

In this paper, we analyzed deep iterative reasoning and identified its reliance on progressively activating pre-trained knowledge. To overcome its challenges, such as ensuring refinement across iterations and determining a stopping criterion, we proposed a breadth reasoning method, which enhances reasoning diversity by modifying input expressions and leveraging self-consistency. Experiments showed that our method effectively expands the reasoning space and outperforms deep iterative reasoning without multiple iterations, highlighting reasoning diversity as a viable alternative to iterative refinement for improving CoT in LLMs.

Several promising directions remain for future work. While modifying pre-defined prompts led to significant performance gains, these modifications remain within the same semantic space. Integrating more effective semantic exploration could further enhance reasoning breadth. 
Additionally, although breadth reasoning successfully circumvents the challenges of deep iterative reasoning, its computational cost cannot be ignored. Future CoT research should focus on efficiently combining depth and breadth reasoning to maximize performance while minimizing computational overhead.

\section*{Limitations}

Although our proposed breadth reasoning approach excels in enhancing reasoning diversity and outperforms deep iterative reasoning in this regard, it still has certain limitations.
First, while it successfully enhances reasoning diversity and outperforms deep iterative reasoning, its reliance on multiple generated reasoning paths increases computational cost, which may not be practical for real-time or resource-constrained applications.

Second, our approach primarily explores reasoning diversity within a fixed semantic space by modifying input expressions. However, this modification may not fully capture deeper semantic variations that could further improve reasoning breadth. Developing techniques to explore broader semantic spaces remains an open challenge.

Lastly, although breadth reasoning circumvents the challenges of iterative refinement, it does not explicitly incorporate mechanisms to ensure reasoning consistency across diverse paths. Future work could investigate hybrid approaches that balance depth and breadth reasoning to further enhance the effectiveness of CoT reasoning in LLMs.

% There remain several promising directions for further exploration and improvement. First, we achieved significant performance gains merely by modifying the phrasing of pre-defined prompts. However, these modifications remain constrained within the same semantic space. Incorporating more effective semantic exploration alongside expression refinement could further expand the breadth of reasoning. Second, while breadth reasoning circumvents the challenges associated with deep iterative reasoning, its increased computational cost cannot be overlooked. Therefore, future research on CoT should focus on effectively integrating the advantages of both depth and breadth reasoning to achieve a more optimal balance.

% 宽度的预定义prompt可以进一步优化，宽度和深度可以结合取长补短

\bibliography{custom}

\newpage

\appendix

\section{Appendix}
\label{sec:appendix}

\subsection{Related work}
\subsubsection{Chain-of-Thought}

Chain-of-thought reasoning (CoT) \cite{kojima2022large,wei2022chain}, a prompting-based approach that guides LLMs to reason step by step, has emerged as a powerful paradigm for enhancing their decision-making capabilities. CoT has been widely applied to various tasks, including mathematical problem-solving \cite{mishra2022lila} and multi-modal reasoning \cite{chen2023measuring,lu2022learn}, due to its improved reasoning performance, interpretability, transparency \cite{wang2023knowledge}, and collaborative capabilities \cite{le2024visualcoder}.

Traditional CoT methods primarily focus on prompt construction. Manual prompting techniques, such as PAL \cite{gao2023pal}, achieve high performance but are costly and difficult to generalize. In contrast, automatic prompting methods, such as Auto-CoT \cite{zhang2022automatic}, offer low-cost, easily transferable solutions but are more error-prone. To balance performance and efficiency, semi-automatic prompting methods, including AutoMate CoT \cite{shum2023automatic} and BoostedPrompt \cite{pitis2023boosted}, have been introduced, making them suitable for real-world applications.

Beyond prompt design, several different approaches address challenges in CoT reasoning. Self-Refine \cite{madaan2024self} and Reflexion \cite{shinn2024reflexion} enhance LLM reliability by effectively mitigating hallucinations and factual inaccuracies through iterative verification and refinement. To better overcome vanilla CoT’s limitations in handling complex questions, L2M \cite{zhou2022least} decompose problems into simpler sub-tasks. In highly knowledge-sensitive tasks, KD-CoT \cite{wang2023knowledge} integrate external knowledge bases to improve accuracy and reduce factual errors. Additionally, RankPrompt \cite{hu2024rankprompt} leverages inherent uncertainty through self-ensemble techniques, allowing LLMs to rank predictions and enhance accuracy.

Driven by the structural limitations of sequential CoT reasoning, recent advancements have introduced more expressive architectures, such as tree- and graph-based structures. Tree structures \cite{yao2024tree,chen2024boosting} facilitate broader exploration and enable backtracking, enhancing the model’s ability to refine its reasoning process. Meanwhile, graph structures \cite{lei2023boosting,besta2024graph} improve sub-problem aggregation and enable self-verification, further strengthening the robustness and coherence of CoT reasoning.

\subsubsection{Long-Chain Thought Reasoning}
The foundational concept of CoT reasoning has been significantly extended through the development of long-chain thought reasoning. Pioneering models such as OpenAI o1 series \cite{O1} have introduced inference-time scaling by lengthening the reasoning process, enabling more sophisticated deliberation. The key advantage of long-chain CoT lies in its ability to break down complex problems into finer-grained steps, fostering deeper analytical reasoning and ultimately leading to more precise and comprehensive solutions.

Beyond the OpenAI o1, several notable models have embraced the long-chain reasoning paradigm. For instance, DeepSeek-R1 \cite{deepseek}, QwQ \cite{qwq}, and Marco-o1 \cite{zhao2024marco} have demonstrated the effectiveness of this approach. These models iteratively refine their reasoning by identifying and correcting errors, simplifying intricate steps, and exploring alternative strategies when necessary, thereby enhancing both the robustness and adaptability of their inference.

Moreover, some models have incorporated verifiable reward mechanisms to refine long-chain CoT generation while addressing challenges such as reward hacking in large-scale reinforcement learning. Specifically, methods that leverage accuracy-driven rewards from ground-truth answers \cite{team2025kimi,tinyzero} help ensure the reliability and consistency of the generated reasoning paths.

The impact of long-chain CoT extends beyond text-based reasoning. For example, Mulberry \cite{yao2024mulberry} has shown that o1-like reasoning principles can be effectively applied to multimodal contexts, expanding the applicability of long-chain CoT across diverse domains. Additionally, recent advancements have focused on improving stepwise coherence and integrating diverse reasoning strategies \cite{team2025kimi}, leading to enhanced model performance and training efficiency.

These developments underscore the transformative potential of long-chain CoT in enabling LLMs to engage in more structured, rigorous, and adaptable problem-solving, paving the way for more advanced and reliable AI reasoning systems.

\subsection{Algorithmic Pseudo-Code}

We provide the pseudo-code of our proposed method (\eg QuestionC-SC) in Algorithm \ref{alg:algorithmname}.

\begin{algorithm}[t]
\caption{Expanding Reasoning Breadth via Context Reformulation}
\label{alg:breadth}
\KwIn{Given question $q$, large language model \textit{LLM}, number of reformulations $N$, self-consistency samples per reformulation $M$}
\KwOut{Final answer $A^*$}

\tcp{Step 1: Generate multiple reformulated prompts}
\textbf{Instruction:} \textit{``Rephrase the sentence to change its wording and structure while maintaining the same meaning. Ensure the core sentence remains unchanged.''}

\For{$i=1$ to $N$}{
    Generate reformulated question $q_i$ by prompting the \textit{LLM} with the instruction and original question $q$;
}

\tcp{Step 2: Perform CoT reasoning with self-consistency sampling}
\For{$i=1$ to $N$}{
    \For{$j=1$ to $M$}{
        Generate reasoning path $R_{i,j}$ using CoT reasoning on $q_i$;
        Obtain prediction $A_{i,j}$ from $R_{i,j}$;
    }
}

\tcp{Step 3: Aggregate predictions through major voting}
Obtain final answer $A^* = \text{Vote}(\{A_{i,j}\})$;

\Return $A^*$;
    \label{alg:algorithmname}
\end{algorithm}

\subsection{Experimental Results of Section \ref{aif}}

\subsubsection{Experimental Setting}

This section outlines the experimental setup, detailing the models, datasets, and the variations introduced under different experimental conditions. Specifically, we design four types of experiments: (i) modify the expression of the given question, (ii) modify the expression of the pre-defined prompt, (iii) perturbations applied to LLMs themselves, and (iv) sample initial reasoning paths from LLMs.

\textbf{Model:} Our experiments using the open-source model Llama-2-7b-chat, released by Meta. Llama-2-7b-chat is an optimized version of Llama-2-7b for conversational use cases. The model weights can be accessed from \url{https://huggingface.co/meta-llama/Llama-2-7b-chat}.

\textbf{Datasets:}
Rather than evaluating the entire dataset, we construct a challenging subset comprising problems that LLMs struggle to solve. Specifically, we select 20 questions from the AddSub dataset and another 20 from the AQuA dataset, forming two separate test sets.
% Additionally, we include two commonsense reasoning datasets, CommonsenseQA and StrategyQA, to assess generalization across different reasoning tasks.

\textbf{Computational Setup}: 
All experiments are conducted on 8 NVIDIA GeForce RTX 3090 GPUs. The top-k parameter is set to 10. Unless otherwise stated, we use ``Let's think step by step.'' as the CoT prompt trigger for all experiments except for those involving variations in prompt expressions. Similarly, except for the variations introduced by LLM sampling, the temperature parameter is set to 1.0 for all other experiments.

\textbf{Variations in Question Expressions}:
To generate different question formulations while preserving their meaning, we use the following instruction:
\textbf{\# Instruction:} \textit{Significantly rephrase the following question to change its wording and structure while maintaining the same meaning and intent. Ensure that the core problem remains unchanged.
}
For each question, this prompt generates 10 reworded versions, including the original, which are used as input for the CoT prompting mechanism.

\begin{table*}[h!]
\centering
  \resizebox{\textwidth}{!}{
\begin{tabular}{lcccccl}
\toprule
Dataset & Answer Format (*1) & \# of samples & Avg \# words (*2) & Data split (filename) & License \\
\midrule
SingleEq          & N & 508  & 27.4  & questions.json            & No License       \\
AddSub            & N & 395  & 31.5  & AddSub.json               & Unspecified      \\
MultiArith        & N & 600  & 31.8  & MultiArith.json           & Unspecified      \\
GSM8K             & N & 1319 & 46.9  & test.jsonl                & MIT License      \\
AQUA          & M & 254  & 51.9  & test.jsonl                & Apache-2.0       \\
SVAMP             & N & 1000 & 31.8  & SVAMP.json                & MIT License      \\
CommonsenseQA     & M & 1221 & 27.8  & dev\_rand\_split.jsonl    & Unspecified      \\
StrategyQA        & Y & 2290 & 9.6   & task.json                & Apache-2.0       \\
LastLetters      & F & 500  & 15.0  & -                        & -                \\
CoinFlip         & Y & 500  & 37.0  & -                        & -                \\
\bottomrule
\end{tabular}}
    % \vspace{-4pt}
\caption{Detailed description of the datasets used in our experiments, carefully 
highlighting their diversity and structure. (\*1) The ``Answer Format" column indicates the type of responses typically expected for each dataset: N represents a numerical answer, M corresponds to selecting one option from multiple possible choices, Y indicates a binary answer (Yes or No), and F stands for free-form answers. (\*2) The ``Avg \# words" column approximately represents the average number of words in the question texts, providing an estimate of their complexity.}
    % \vspace{-12pt}
    \label{tab:detaileddataset}
\end{table*}

\textbf{Variations in Prompt Expressions:}
To evaluate the impact of different prompt formulations on reasoning, we use the following 10 alternative prompts to induce LLMs to generate reasoning chains:
\textit{(1) Here are the steps we can follow to achieve our goal.
(2) How about we break it down into smaller parts?
(3) How would you like to approach this situation in a methodical manner?
(4) Let's break down the problem into smaller parts and tackle each one separately.
(5) Here are some thoughtful steps to consider.
(6) Let's take it one step at a time.
(7) Let's take a thoughtful approach.
(8) In a systematic approach, let's work through the following stages.
(9) In a systematic manner, let us consider each detail.
(10) Let's think step by step.}
Each of these prompts generates a distinct reasoning path, allowing us to analyze the sensitivity of the model to different phrasing styles.

\textbf{Variations Introduced by Perturbations to LLMs:}
To introduce controlled perturbations into the LLM, we replace its final linear layer with a noisy linear layer, where both the weights and biases are perturbed. The noise is sampled from a normal distribution, with a standard deviation ranging from 0.010 to 0.020 in increments of 0.001, resulting in 10 distinct perturbed models. Each perturbed model generates a unique reasoning path, which is subsequently processed by the standard LLaMA-2-7B-Chat to obtain the final results.

\textbf{Variations Introduced by LLM Sampling:}
To assess the impact of sampling variability, we set the temperature parameter to 0.8 and the number of samples to 10. This setup generates 10 distinct inference paths, enabling an analysis of how sampling randomness affects reasoning outcomes.

% \begin{table*}[htbp]
% \centering
% % \resizebox{\columnwidth}{!}{ 
% % \tiny
% \begin{tabular}{lccc}
% \hline
% \textbf{Prompt Type} & Iterations 1 \& 2 &  Iterations 2 \& 3 & Avg. \\ \hline
% General Prompt & 0.44 & 0.32 & 0.38 \\
% Our Prompt $p^{\ast}$ & 0.28 & 0.29 & 0.29 \\
% \hline
% $\Delta$ & -0.16 & -0.03 & -0.10\\
% $\Delta$\% & \-36.4\% & -9.4\% & -26.3\% \\
% \hline
% \end{tabular}
% % }
% \caption{The reasoning similarity between new and previous iterations guided by general prompt and our $p^{\ast}$ on the AQuA dataset.}
% \label{tab:similarity}
% \end{table*}

\subsubsection{Main Results}

Table \ref{tab:entropy} presents the information entropy values of predictions generated under different influencing factors, including variations in the question phrasing, prompt formulation, model-intrinsic perturbations, and sampling strategies. The entropy metric quantifies the diversity of reasoning paths, where a higher entropy value indicates a greater exploration of different reasoning trajectories.

Notably, LLM-intrinsic perturbations yield the least entropy, suggesting that the model’s inherent generation process alone is insufficient to significantly diversify reasoning pathways. This result highlights LLMs’ strong robustness, where minor perturbations fail to introduce substantial variations in reasoning paths, leading to consistent predictions even under slight modifications. Sampling strategies exhibit moderate entropy values, indicating that while sampling can introduce some variations in the reasoning process, it remains constrained by the overarching context and prior model biases.

Additionally, we observe that the information entropy values are relatively consistent across datasets, with AQuA and AddSub showing similar trends. The average entropy values reinforce the conclusion that prompt modifications are the most effective in enhancing reasoning diversity, while LLM perturbations alone are the least impactful.

These findings provide empirical evidence for the crucial role of context reconfiguration in diversifying reasoning paths, suggesting that future work on enhancing reasoning diversity should prioritize context manipulation strategies over simple perturbations or sampling techniques.

% From the results, we observe that modifying the prompt expression induces the highest entropy across datasets, followed by direct question perturbations. This finding aligns with our analysis in the main text, where we highlight the strong contextual dependence of LLMs. By altering the prompt or rephrasing the question, the model perceives a different context, leading to more diverse reasoning paths. In contrast, sampling-based variations exhibit lower entropy, as they operate within the same contextual framework and, therefore, provide more constrained diversity. Notably, LLM-intrinsic perturbations yield the least entropy, suggesting that the model’s inherent generation process alone is insufficient to significantly diversify reasoning pathways.

% These findings reinforce the importance of context modulation in enhancing reasoning diversity. While self-consistency mechanisms contribute to some extent, structural modifications to prompts and questions prove to be the most effective in broadening the spectrum of reasoning paths.

\begin{table}[htbp]
  \centering
  \small
  % \resizebox{\textwidth}{!}{
  \begin{tabular}{lcccc}
    \toprule
    \textbf{Datasets} & Question & Prompt & LLMs & Sampling \\
    \midrule
    AQuA & 1.61 & 2.06 & 1.23 & 1.52 \\
    AddSub & 1.60 & 2.04 & 1.26 & 1.54 \\
    \hline
    \textbf{Avg.} & 1.61 & 2.05 & 1.24 & 1.53 \\
    \bottomrule
  \end{tabular}
  % \vspace{-8pt}
  \caption{Information entropy of predictions on a subset of problems that LLMs struggle to solve, showing the impact of different factors on reasoning path diversity.}
  \label{tab:entropy}
  % \vspace{-8pt}
\end{table}

\subsection{Experimental Demos of Deep Iterative Reasoning and Breadth Reasoning}

\begin{table*}[ht]
\scriptsize
    \centering
    \begin{tabular}{p{2.7cm} | p{5.7cm} | p{5.4cm} | p{1.4cm} }
        \hline
        \textbf{Question} & \textbf{Iteration 1} & \textbf{Iteration 2} & \textbf{Output}  \\
        \midrule
        \textit{Alex and Jacob works at a toy shop that make toys. Alex takes 7 hours to make a toy, and Jacob takes 9 hours to make a toy. During a month, both of them makes 35 toys in total. If both of them have worked for almost similar number of hours how many toys have been prepared by Jacob? Answer Choices: (A) 15 (B) 16 (C) 17 (D) 18 (E) 19}
         & \textbf{\textit{Let's think step by step}}.
         Step 1: Calculate the combined work rate of Alex and Jacob
Alex can make 1 toy in 7 hours, so his work rate is 1 toy / 7 hours = 1/7 toys per hour.
Jacob can make 1 toy in 9 hours, so his work rate is 1 toy / 9 hours = 1/9 toys per hour.
The combined work rate of Alex and Jacob is (1/7 + 1/9) toys per hour.

Step 2: Calculate the total number of hours worked in a month
Let the total number of hours worked be x.
Since both Alex and Jacob have worked for almost similar hours, we can assume Alex worked for hx hours and Jacob worked for (1-h)x hours, where h is a fraction close to 0.5.

Step 3: Set up the equation based on the relationship between work rate, time, and total toys
Total toys made by Alex = (1/7) * hx
Total toys made by Jacob = (1/9) * (1-h)x
Total toys made in a month = 35 toys
Therefore, the equation is: (1/7) * hx + (1/9) * (1-h)x = 35

Step 4: Solve the equation to find the value of h and number of toys made by Jacob
Solving the equation (1/7) * hx + (1/9) * (1-h)x = 35 will give you the value of h.
Once you find the value of h, you can calculate the number of toys made by Jacob, which is (1/9) * (1-h)x.

Step 5: Check the answer choices
After calculating the number of toys made by Jacob, check which of the answer choices matches the result you obtained. This will give you the final answer.
Let's start with step 1 and work through the problem together.

{\cm Therefore, among A through E, the answer is (B) 16}.
& \textbf{\textit{Let's think step by step}}.
This problem can be solved using a common formula that involves the rates at which Alex and Jacob work. The formula is:

Rate = Work / Time

From the information given, we know that Alex takes 7 hours to make a toy, and Jacob takes 9 hours to make a toy. So, we can calculate their rates as follows:

Alex's rate = 1 toy / 7 hours
Jacob's rate = 1 toy / 9 hours

Now, it is given that both of them together make 35 toys in total and have worked for similar hours. Let's assume that they both worked for the same number of hours, denoted as x.

So, we can set up an equation based on the rates to solve for the number of toys made by Jacob:

Alex's work + Jacob's work = 35 toys
(x / 7) + (x / 9) = 35
9x + 7x = 35 * 63
16x = 35 * 63
x = 35 * 63 / 16
x = 138.75 hours

Since Jacob works at a rate of 1 toy per 9 hours, we can calculate the number of toys made by Jacob:

Jacob's work = 138.75 / 9
Jacob's work = 15.416...

Rounding to the nearest whole number, Jacob made 15 toys.

{\cmth Therefore, the correct answer is (A) 15}.

         & \makecell{(A) 15 \\ {\cmth Answer: (A)}}   \\
         \midrule
        \textit{A rectangular solid, 3 x 4 x 15, is inscribed in a sphere, so that all eight of its vertices are on the sphere. What is the diameter of the sphere? Answer Choices: (A)  13.3542 (B)  15.8113 (C)  18.3451 (D)  19.5667 (E)  20.8888}
         & \textbf{\textit{Let's think step by step}}.
         Step 1: Find the center of the sphere.
Since the solid is inscribed in the sphere, the center of the sphere is at the center of the rectangular solid. 
The center of the rectangular solid is at the midpoint of the diagonal connecting two opposite vertices of the solid.

Step 2: Find the length of the diagonal of the rectangle.
The length of the diagonal can be found using the Pythagorean theorem.
Using the sides of the rectangle (3, 4, and 15), we can find the length of the diagonal.

Applying the Pythagorean theorem, the diagonal length is:
\( \sqrt{3^2 + 4^2 + 15^2} = \sqrt{9 + 16 + 225} = \sqrt{250} = 5\sqrt{10} \)

Step 3: The diameter of the sphere is equal to the length of the diagonal of the rectangle.
The diameter of the sphere is equal to the length of the diagonal of the rectangle, which is \( 5\sqrt{10} \).

Therefore, the diameter of the sphere is \( 2 \times 5\sqrt{10} = 10\sqrt{10} \approx 31.6228 \).

Among the answer choices provided, the closest value is (E) 20.8888.

{\cm Therefore, among A through E, the answer is (E) 20.8888}.
         & \textbf{\textit{Let's think step by step}}.
         First, let's calculate the length of the diagonal of the rectangular solid. We can use the Pythagorean theorem for this.

The diagonal length \( d \) of the rectangular solid with sides of lengths \( a \), \( b \), and \( c \) can be found using the formula:
\[
d = \sqrt{a^2 + b^2 + c^2}
\]

In this case, the sides of the rectangular solid are 3, 4, and 15:
\[
d = \sqrt{3^2 + 4^2 + 15^2} = \sqrt{250} \approx 15.8114
\]

Therefore, the length of the diagonal of the rectangular solid is approximately 15.8114.

Since the diagonal of the rectangular solid is also the diameter of the inscribed sphere, the diameter of the sphere is approximately 15.8114.

Among the answer choices provided, the closest value to 15.8114 is (B) 15.8113.

So, the answer is (B) 15.8113.

{\cmth Therefore, among A through E, the answer is (B) 15.8113}.

         & \makecell{(B) 15.8113 \\{\cmth Answer: (B)}}  \\
        \bottomrule
    \end{tabular}
    \caption{Problems solvable by both depth and breadth reasoning \textbf{(illustrating depth reasoning here)}.}
\end{table*}

\begin{table*}[ht]
\scriptsize
    \centering
    \begin{tabular}{p{2.7cm} | p{11.1cm} | p{1.4cm} }
        \hline
        \textbf{Question} & \textbf{Reasoning Process} & \textbf{Output}  \\
        \midrule
        \textit{Alex and Jacob works at a toy shop that make toys. Alex takes 7 hours to make a toy, and Jacob takes 9 hours to make a toy. During a month, both of them makes 35 toys in total. If both of them have worked for almost similar number of hours how many toys have been prepared by Jacob? Answer Choices: (A) 15 (B) 16 (C) 17 (D) 18 (E) 19}
         & \textbf{Path 1:} \textbf{\textit{Let's think step by step}}.
1. Let's denote the number of toys Alex makes as \( x \) and the number of toys Jacob makes as \( y \).
2. We know that Alex takes 7 hours to make one toy and Jacob takes 9 hours to make one toy. So, their work rates are \(\frac{1}{7}\) and \(\frac{1}{9}\) toys per hour, respectively.
3. Since they make 35 toys together, we can set up the following equation: 
\[ x\left(\frac{1}{7}\right) + y\left(\frac{1}{9}\right) = 35 \]
4. From the question, we also know that they have worked for almost similar number of hours. Therefore, we can assume that \(x\) and \(y\) are close in value. Let's assume \(x = y = z\).
5. Substituting this assumption into the equation, we get:
\[ z\left(\frac{1}{7} + \frac{1}{9}\right) = 35 \]
\[ z\left(\frac{16}{63}\right) = 35 \]
6. Solving for \(z\), we get:
\[ z = \frac{35\times 63}{16} \]
\[ z = 138.75 \]
7. Since we cannot have a fraction of a toy, we know that the closest whole number to 138.75 is 139. Therefore, both Alex and Jacob made approximately 139 toys each.
8. Since the question asks for the number of toys prepared by Jacob, the answer is approximately 139, which is closest to choice (C) 17.

{\cm Therefore, among A through E, the answer is (C) 17}.

\textbf{Path 2:} \textbf{\textit{Let's approach this systematically, one step at a time}}.
Let's denote the number of toys Alex made as x, and the number of toys Jacob made as y. 

Given that Alex takes 7 hours to make a toy, and Jacob takes 9 hours to make a toy, we can set up the following equation based on the time they worked:
7x = 9y

Since both of them made a total of 35 toys, we have another equation:
x + y = 35

We can solve these two equations simultaneously to find the values of x and y.

First, let's solve for y in terms of x from the first equation:
y = (7x) / 9

Now substitute y = (7x) / 9 into the second equation:
x + (7x) / 9 = 35
9x + 7x = 315
16x = 315
x = 19.6875

Since we cannot make a fraction of a toy, we round down to the nearest whole number. Therefore, Alex made 19 toys. 

Now we can calculate how many toys Jacob made:
y = (7 * 19) / 9
y = 133 / 9
y = 14.777

Rounding down to the nearest whole number, Jacob made 14 toys during the month.

Therefore, the number of toys prepared by Jacob is 14. 

{\cmth The correct answer choice is (A) 15}.

\textbf{Path 3:} \textbf{\textit{Let's break it down systematically}}.
Let's denote the number of toys Alex made as x and the number of toys Jacob made as y.

We are given that Alex takes 7 hours to make one toy and Jacob takes 9 hours to make one toy. Therefore, the total number of hours Alex worked would be 7x and the total number of hours Jacob worked would be 9y.

From the given information, we know that the total number of toys made by both Alex and Jacob is 35:
x + y = 35

Also, we know that both Alex and Jacob worked for approximately the same amount of time:
7x = 9y

Let's solve these two equations simultaneously to find the number of toys prepared by Jacob:
7x = 9y
7(35 - y) = 9y
245 - 7y = 9y
245 = 16y
y = 15.31 (approx.)

Since y represents the number of toys made by Jacob, we can conclude that Jacob made approximately 15 toys. Therefore, the correct answer is (A) 15.

{\cmth Therefore, among A through E, the answer is (A) 15}.

\textbf{Path 4:}  \textbf{\textit{Let's break it down systematically}}.
Let's denote the number of toys Alex made as x and the number of toys Jacob made as y.

We are given that Alex takes 7 hours to make one toy and Jacob takes 9 hours to make one toy. Therefore, the total number of hours Alex worked would be 7x and the total number of hours Jacob worked would be 9y.

From the given information, we know that the total number of toys made by both Alex and Jacob is 35:
x + y = 35

Also, we know that both Alex and Jacob worked for approximately the same amount of time:
7x = 9y

Let's solve these two equations simultaneously to find the number of toys prepared by Jacob:
7x = 9y
7(35 - y) = 9y
245 - 7y = 9y
245 = 16y
y = 15.31 (approx.)

Since y represents the number of toys made by Jacob, we can conclude that Jacob made approximately 15 toys. 

{\cmth Therefore, the correct answer is (A) 15}.

         & \makecell{(A) 15 \\ {\cmth Answer: (A)}}   \\  
        
        \bottomrule
    \end{tabular}
    \caption{Problems solvable by both depth and breadth reasoning \textbf{(illustrating breadth reasoning here)}.}
\end{table*}

\begin{table*}[ht]
\scriptsize
    \centering
    \begin{tabular}{p{2.7cm} | p{11.1cm} | p{1.4cm} }
        \hline
        \textbf{Question} & \textbf{Reasoning Process} & \textbf{Output}  \\
        \midrule
        \textit{A rectangular solid, 3 x 4 x 15, is inscribed in a sphere, so that all eight of its vertices are on the sphere. What is the diameter of the sphere? Answer Choices: (A)  13.3542 (B) 15.8113 (C)  18.3451 (D)  19.5667 (E)  20.8888
A: Let's think step by step.}
         & \textbf{Path 1:} \textbf{\textit{Let's break this problem down into steps}}:

**Step 1: Find the diagonal of the rectangular solid.**  
The diagonal of the rectangular solid is calculated using the Pythagorean theorem in three dimensions:

\[
\text{Diagonal} = \sqrt{3^2 + 4^2 + 15^2}
\]

\[
= \sqrt{9 + 16 + 225}
\]

\[
= \sqrt{250}
\]

\[
\approx 15.8114
\]

**Step 2: The diameter of the sphere is equal to the diagonal of the rectangular solid.**  
Therefore, the diameter of the sphere is approximately **15.8114** units.

So, the closest answer choice is **15.8113**, which corresponds to option **(B)**.

**Revisiting the diagonal calculation for confirmation:**  
First, let's find the length of the diagonal of the rectangular solid. This can be done using the Pythagorean theorem.

The diagonal of the rectangular solid is the hypotenuse of a right triangle with legs of 3, 4, and the diagonal as the hypotenuse.

Using the Pythagorean theorem:

\[
d^2 = 3^2 + 4^2
\]

\[
d^2 = 9 + 16 = 25
\]

\[
d = \sqrt{25} = 5
\]

So, the diagonal of this smaller rectangular face is **5 units**.

Next, considering the full three-dimensional diagonal:

Since the diagonal of the sphere is equal to the diagonal of the rectangular solid (passing through the center of the sphere), we compute:

\[
d = \sqrt{3^2 + 4^2 + 15^2}
\]

\[
= \sqrt{9 + 16 + 225}
\]

\[
= \sqrt{250}
\]

\[
\approx 15.8114
\]

Thus, the correct diagonal length of the rectangular solid is approximately **15.8114** units.

Since the closest answer choice is **15.8113**, {\cmth the correct answer among the options provided is **(B) 15.8113**}.

\textbf{Path 2:} \textbf{\textit{Let's think step by step}}.

**Step 1:** Find the diagonal of the rectangular solid using the Pythagorean theorem.  
The diagonal of the rectangular solid is the diameter of the sphere when inscribed.

**Step 2:** The diagonal of the rectangular solid can be found by calculating the square root of the sum of the squares of the dimensions:

\[
\sqrt{3^2 + 4^2 + 15^2} = \sqrt{9 + 16 + 225} = \sqrt{250}
\]

**Step 3:** Simplify the square root of 250:

\[
\sqrt{250} = \sqrt{25 \times 10} = \sqrt{25} \times \sqrt{10} = 5\sqrt{10}
\]

Therefore, the diameter of the sphere is \( 5\sqrt{10} \). Let's calculate the approximate value.

{\cmth Thus, among choices A through E, the answer is **(B) 15.8113**}.

\textbf{Path 3:} \textbf{\textit{Let's break it down systematically}}.

Sure! Let's first find the diagonal of the rectangular solid, which will be the diameter of the sphere when inscribed.

The diagonal of the rectangular solid can be found using the Pythagorean theorem in three dimensions:

\[
d^2 = 3^2 + 4^2 + 15^2
\]

\[
d^2 = 9 + 16 + 225
\]

\[
d^2 = 250
\]

\[
d = \sqrt{250}
\]

\[
d \approx 15.8114
\]

Therefore, the diameter of the sphere, when inscribed in the rectangular solid, {\cmth is approximately **15.8114**.  
So the closest answer choice is **(B) 15.8113**}.

         & \makecell{(B) 15.8113 \\ {\cmth Answer: (B)}}   \\  
        
        \bottomrule
    \end{tabular}
    \caption{Problems solvable by both depth and breadth reasoning \textbf{(illustrating breadth reasoning here)}.}
\end{table*}

\begin{table*}[ht]
\scriptsize
    \centering
    \begin{tabular}{p{2.7cm} | p{5.7cm} | p{5.4cm} | p{1.4cm} }
        \hline
        \textbf{Question} & \textbf{Iteration 1} & \textbf{Iteration 2} & \textbf{Output}  \\
        \midrule
        \textit{A car is being driven, in a straight line and at a uniform speed, towards the base of a vertical tower. The top of the tower is observed from the car and, in the process, it takes 10 minutes for the angle of elevation to change from 45° to 60°. After how much more time will this car reach the base of the tower?
        Answer Choices: (A) \( 5 (\sqrt{3} + 1) \)  
(B) \( 6 (\sqrt{3} + \sqrt{2}) \)  
(C) \( 7 (\sqrt{3} - 1) \)  
(D) \( 8 (\sqrt{3} - 2) \)  
(E) None of these}
         & \textbf{\textit{Let's think step by step}}.
Great job breaking down the problem step by step! If you have any more questions or need further assistance, feel free to ask.

Therefore, among A through E, the answer is The answer is (E) None of these. Let's continue solving the problem.

The change in angle of elevation from 45° to 60° corresponds to an angle change of 15°. This change happened in 10 minutes.

To find out how much further time it will take for the car to reach the base of the tower, we need to determine the time taken for an angle change of 30° (from 60° to 90°) as the car will be at the base when the angle of elevation is 90°.

Since the time taken is directly proportional to the angle change, we can set up a proportion: 

10 minutes is to 15° as X minutes is to 30°.

This gives us the equation: 10/15 = X/30

Solving for X, we get: X = 20 minutes

{\cm Therefore, the car will take an additional 20 minutes to reach the base of the tower from the point where the angle of elevation is 60°}.

& \textbf{\textit{Let's think step by step}}.

**Step 1:** Let's start by setting up the situation. The car is approaching the tower, and it takes 10 minutes for the angle of elevation to change from \( 45^\circ \) to \( 60^\circ \). This means that the car is moving towards the tower at a constant speed. Let's denote the distance between the car and the tower as \( x \) units.

**Step 2:** At the initial position, when the angle of elevation is \( 45^\circ \), we can construct a right triangle. The base of the triangle is \( x \) units (distance between the car and the tower), the height is the height of the tower (which we don't know), and the angle is \( 45^\circ \).

**Step 3:** When the angle of elevation changes to \( 60^\circ \) after 10 minutes, we have a new right triangle with the same height of the tower and the base now reduced to \( y \) units (distance traveled by the car). From trigonometry, we can set up the following equations:

\[
\tan 45^\circ = \frac{\text{height}}{x}
\]

\[
\tan 60^\circ = \frac{\text{height}}{y}
\]

**Step 4:** From the given information, we know that \( \tan 45^\circ = 1 \) and \( \tan 60^\circ = \sqrt{3} \). Therefore, we have:

\[
1 = \frac{\text{height}}{x}
\]

\[
\sqrt{3} = \frac{\text{height}}{y}
\]

**Step 5:** By solving these two equations simultaneously, we can find the ratio of \( x \) to \( y \). This will help us determine how much more time the car needs to reach the base of the tower. Let's calculate it.

{\cm Therefore, among A through E, the answer is **Option D: \( 8 (\sqrt{3} - 2) \)**}.
 
         & \makecell{{\cm (D) \( 8 (\sqrt{3} - 2) \)} \\ {\cmth Answer: (A)}}   \\
         \midrule
        \textit{Find the total no. of distinct bike no.'s that can beformed using 2 letters followed by 2 no.'s. How many letters need to be distinct? Answer Choices: (A) 74453 (B) 64543 (C) 74325 (D) 65000 (E) 97656}
         & \textbf{\textit{Let's think step by step}}.
A: Let's think step by step.
Let's break down the problem step by step:

Step 1: Select 2 distinct letters
There are 26 letters in the English alphabet. To choose 2 distinct letters, we use the combination formula: 26C2 = 26! / (2! * (26-2)!) = 325 possibilities.

Step 2: Select 2 distinct numbers
There are 10 digits from 0 to 9. To select 2 distinct numbers, we use the combination formula: 10C2 = 10! / (2! * (10-2)!) = 45 possibilities.

Step 3: Multiply the possibilities for letters and numbers
To find the total number of distinct bike numbers, we multiply the possibilities for selecting the letters and the numbers: 325 x 45 = 14,625.

Therefore, the total number of distinct bike numbers that can be formed using 2 letters followed by 2 numbers is 14,625.

Among the answer choices provided, the closest one to 14,625 is (A) 74453.

So, the correct answer is (A) 74453.

{\cm Therefore, among A through E, the answer is (A) 74453}
         & \textbf{\textit{Let's break down the problem step by step}}:

Step 1: Select 2 distinct letters
There are 26 letters in the English alphabet. To choose 2 distinct letters, we use the combination formula: 26C2 = 26! / (2! * (26-2)!) = 325 possibilities.

Step 2: Select 2 distinct numbers
There are 10 digits from 0 to 9. To select 2 distinct numbers, we use the combination formula: 10C2 = 10! / (2! * (10-2)!) = 45 possibilities.

Step 3: Multiply the possibilities for letters and numbers
To find the total number of distinct bike numbers, we multiply the possibilities for selecting the letters and the numbers: 325 x 45 = 14,625.

Therefore, the total number of distinct bike numbers that can be formed using 2 letters followed by 2 numbers is 14,625.

Among the answer choices provided, the closest one to 14,625 is (A) 74453.

{\cm So, the correct answer is (A) 74453}.

         & \makecell{{\cm (A) 74453} \\{\cmth (Answer: (D)}}  \\
        \bottomrule
    \end{tabular}
\caption{Problems solvable by breadth reasoning but not depth reasoning \textbf{(illustrating depth reasoning here)}.}

\end{table*}

\begin{table*}[ht]
\scriptsize
    \centering
    \begin{tabular}{p{2.7cm} | p{11.1cm} | p{1.4cm} }
        \hline
        \textbf{Question} & \textbf{Reasoning Process} & \textbf{Output}  \\
        \midrule
        \textit{A car is being driven, in a straight line and at a uniform speed, towards the base of a vertical tower. The top of the tower is observed from the car and, in the process, it takes 10 minutes for the angle of elevation to change from 45° to 60°. After how much more time will this car reach the base of the tower?
        Answer Choices: (A) \( 5 (\sqrt{3} + 1) \)  
(B) \( 6 (\sqrt{3} + \sqrt{2}) \)  
(C) \( 7 (\sqrt{3} - 1) \)  
(D) \( 8 (\sqrt{3} - 2) \)  
(E) None of these}. &

\textbf{\textit{Let's break it down step by step to solve this problem}}:

1. Let the distance between the car and the base of the tower be \( x \) units.
2. From the triangle formed by the car, the top of the tower, and the base of the tower, we can see that the height of the tower is:
   \[
   x \cdot \tan 45^\circ = x.
   \]
3. When the angle changes from \( 45^\circ \) to \( 60^\circ \), the height of the tower is:
   \[
   x \cdot \tan 60^\circ = x \sqrt{3}.
   \]
4. The difference in height observed by the car during this angle change is:
   \[
   x (\sqrt{3} - 1).
   \]
5. Since the car is moving towards the tower at a uniform speed, the rate at which the observed height changes is constant.
6. It took 10 minutes for the angle of elevation to change from \( 45^\circ \) to \( 60^\circ \), so the car traveled \( x (\sqrt{3} - 1) \) in 10 minutes.
7. The total distance the car needs to cover to reach the base of the tower is \( x \), so the time it will take to cover the remaining distance is:
   \[
   \frac{x}{x (\sqrt{3} - 1)} \times 10 \text{ minutes}.
   \]
8. Simplifying, we get:
   \[
   \frac{10}{\sqrt{3} - 1} \text{ minutes}.
   \]
9. Rationalizing the denominator:
   \[
   \frac{10 (\sqrt{3} + 1)}{2} = 5 (\sqrt{3} + 1) \text{ minutes}.
   \]

Therefore, the answer is **(A) \( 5 (\sqrt{3} + 1) \)**.

{\cmth Thus, among choices A through E, the correct answer is **(A) \( 5 (\sqrt{3} + 1) \)**}.

\textbf{Path 2:} \textbf{\textit{Let's break down the problem step by step}}:

1. When the angle of elevation changes from \( 45^\circ \) to \( 60^\circ \), a right triangle is formed between the car, the top of the tower, and the base of the tower.

2. Let \( x \) be the distance traveled by the car during the 10 minutes it took for the angle of elevation to change. This distance also represents the horizontal distance between the car and the base of the tower.

3. Since the car is moving at a uniform speed, the ratio of the vertical distance and horizontal distance in the right triangle remains constant.

4. Using trigonometry, we can set up equations involving the tangent of the angles:

   \[
   \tan 45^\circ = \frac{h}{x}, \quad \tan 60^\circ = \frac{h - d}{x}
   \]

   where \( h \) is the height of the tower and \( d \) is the distance of the car from the base of the tower.

5. Simplifying and solving the equations, we find:

   \[
   h = x, \quad h - d = x\sqrt{3}.
   \]

6. Since it took 10 minutes for the angle of elevation to change, we can calculate the time to cover distance \( x \) using the speed and distance relation:

   \[
   \text{speed} = \frac{\text{distance}}{\text{time}}.
   \]

7. Speed can be considered as \( \frac{x}{10} \), and the total time taken to reach the base of the tower can be expressed as:

   \[
   \frac{x}{10} + d.
   \]

8. By substituting the value of \( d = h - x = x\sqrt{3} - x \), we can simplify the expression to find the total time taken for the car to reach the base of the tower.

{\cmth Therefore, among choices A through E, the answer is **(A) \( 5 (\sqrt{3} + 1) \)**}.

         & \makecell{(A) \( 5 (\sqrt{3} + 1) \) \\ {\cmth Answer: (A)}}   \\  
        
        \bottomrule
    \end{tabular}
\caption{Problems solvable by breadth reasoning but not depth reasoning \textbf{(illustrating breadth reasoning here)}.}
\end{table*}

\begin{table*}[ht]
\scriptsize
    \centering
    \begin{tabular}{p{2.7cm} | p{11.1cm} | p{1.4cm} }
        \hline
        \textbf{Question} & \textbf{Reasoning Process} & \textbf{Output}  \\
        \midrule
        \textit{Find the total no. of distinct bike no.'s that can beformed using 2 letters followed by 2 no.'s. How many letters need to be distinct? Answer Choices: (A) 74453 (B) 64543 (C) 74325 (D) 65000 (E) 97656}. &

\textbf{Path 1:}
\textbf{\textit{Let's break it down systematically}}.
To find the total number of distinct bike numbers that can be formed using 2 letters followed by 2 numbers, we need to consider the number of possibilities for each part.

For the first two letters, there are 26 choices for each position (A-Z). Since we want the letters to be distinct, the number of ways to choose 2 distinct letters from 26 is calculated as 26P2 = 26! / (26-2)!. This equals 26 * 25 = 650.

For the last two numbers, there are 10 choices for each position (0-9). So, the number of ways to choose 2 distinct numbers from 10 is calculated as 10P2 = 10! / (10-2)! = 10 * 9 = 90.

Therefore, the total number of distinct bike numbers that can be formed is found by multiplying the number of ways to choose 2 distinct letters with the number of ways to choose 2 distinct numbers: 650 * 90 = 58500.

Unfortunately, the correct answer is not listed among the provided choices.

Therefore, among A through E, the answer is 
Sure, let's break down the problem step by step.

We need to form a bike number using 2 letters followed by 2 numbers.

1. Determine the number of ways to select 2 letters:
There are 26 alphabets (A-Z) to choose from. We need to select 2 distinct letters, so we can choose the first letter in 26 ways and the second letter in 25 ways (since it must be different from the first). So, the number of ways to select 2 distinct letters is 26 * 25.

2. Determine the number of ways to select 2 numbers:
There are 10 digits (0-9) to choose from. We can select any of these digits for each of the two numbers. So, the number of ways to select 2 numbers is 10 * 10.

3. Calculate the total number of distinct bike numbers:
To find the total number of distinct bike numbers, we multiply the number of ways to choose the letters by the number of ways to choose the numbers:
26 * 25 * 10 * 10 = 65,000.

{\cmth Therefore, the correct answer is (D) 65,000}.

\textbf{Path 2:} \textbf{\textit{Let's consider things one step at a time}}.
First, let's determine the number of distinct bike numbers that can be formed using 2 letters followed by 2 numbers.

For the 2 letters, since they need to be distinct, we have 26 choices for the first letter and 25 choices for the second letter (since the second letter cannot be the same as the first one).

For the 2 numbers, we have 10 choices for each number (0-9).

So, the total number of distinct bike numbers can be formed is:
26 * 25 * 10 * 10 = 65,000

{\cmth Therefore, the correct answer choice is (D) 65,000}.

\textbf{Path 3:} \textbf{\textit{Let's approach this systematically, one step at a time}}.
Sure! To find the total number of distinct bike numbers that can be formed using 2 letters followed by 2 numbers, we need to consider the following:

1. Number of distinct letters to choose from: Given that the letters need to be distinct, we need to select 2 different letters from the alphabet. There are 26 alphabet letters to choose from, so the number of ways to choose 2 distinct letters is 26 * 25 = 650.

2. Number of distinct numbers to choose from: Similarly, there are 10 distinct numbers (0-9) to choose from for each of the two number positions. So, the number of ways to choose 2 distinct numbers is 10 * 9 = 90.

3. Multiply the number of ways to choose distinct letters and numbers to get the total number of distinct bike numbers: 650 * 90 = 58500.

Therefore, the correct answer choice is not listed among the answer choices provided. The closest one is (D) 65000, but the accurate calculation is 58500.

Therefore, among A through E, the answer is 
To find the total number of distinct bike numbers that can be formed using 2 letters followed by 2 numbers, we first need to understand the requirements.

1. **Selecting 2 distinct letters**: There are 26 letters in the alphabet, so we can select the first letter in 26 ways and the second letter in 25 ways (since the letters should be distinct). This gives us \(26 \times 25 = 650\) ways to choose 2 distinct letters.

2. **Selecting 2 numbers**: For each number, we have 10 choices (0-9). So, for 2 numbers, we have \(10 \times 10 = 100\) ways to choose 2 numbers.

3. **Combining letters and numbers**: Once we have selected the letters and the numbers, we can combine them in any order. So, the total number of ways to form a bike number with 2 letters followed by 2 numbers is \(650 \times 100 = 65000\).

{\cmth Therefore, the correct answer is (D) 65000}.

\textbf{Path 4:}
\textbf{\textit{Let's think step by step}}.
Sure, let's break it down step by step.

1. **Choosing 2 Letters**: We have 26 letters from A to Z. We need to choose 2 letters which can be repeated. So the number of ways to select 2 letters is 26 * 26 = 676.

2. **Choosing 2 Numbers**: We have 10 digits from 0 to 9. We need to choose 2 numbers which can be repeated. So the number of ways to select 2 numbers is 10 * 10 = 100.

3. **Total Number of Distinct Bike Numbers**: To find the total number of distinct bike numbers, we multiply the number of ways to select letters by the number of ways to select numbers: 676 * 100 = 67,600.

Therefore, the correct number of distinct bike numbers that can be formed using 2 letters followed by 2 numbers is 67,600.

So, the answer choice closest to this value is (E) 97656.

{\cm Therefore, among A through E, the answer is (E) 97656}.

         & \makecell{(D) 65000 \\ {\cmth Answer: (D)}}   \\  
        
        \bottomrule
    \end{tabular}
\caption{Problems solvable by breadth reasoning but not depth reasoning \textbf{(illustrating breadth reasoning here)}.}
\end{table*}

\begin{table*}[ht]
\scriptsize
    \centering
    \begin{tabular}{p{2.7cm} | p{5.7cm} | p{5.4cm} | p{1.4cm} }
        \hline
        \textbf{Question} & \textbf{Iteration 1} & \textbf{Iteration 2} & \textbf{Output}  \\
        \midrule
        \textit{Of the 200 students in a school, at least 45\% attended the prom night and at least 35\% took part in the debating session. What is the maximum number of students who could have neither attended the prom night nor the debating session? Answer Choices: (A) 27 (B) 81 (C) 90 (D) 99 (E) 110}
         & \textbf{\textit{Let's think step by step}}.
Sure! Let's break it down step by step:

1. We know that at least 45\% of 200 students attended the prom night, so that means a minimum of \(0.45 \times 200 = 90\) students attended the prom.

2. Also, at least 35\% of 200 students took part in the debating session, so that means a minimum of \(0.35 \times 200 = 70\) students participated in the debating session.

3. The maximum number of students who could have participated in both the prom night and the debating session is the minimum of the number of students who attended the prom and those who participated in the debating session: \(\min(90, 70) = 70\).

4. Therefore, the maximum number of students who could have attended either the prom night or the debating session is \(90 + 70 = 160\) students. This means that at most \(200 - 160 = 40\) students could have neither attended the prom nor the debating session.

5. Since the question asks for the maximum number of students who could have neither attended the prom night nor the debating session, the answer is 40, which is closest to (A) 27.

Therefore, among the answer choices given, the correct choice is (A) 27.

{\cm Therefore, among A through E, the answer is (A) 27}.
& \textbf{\textit{Let's think step by step}}.
Sure! Let's break it down step by step.

1. In a group of 200 students, at least 45\% attended the prom night.
- 45\% of 200 is 0.45 x 200 = 90 students attended the prom.

2. At least 35\% of the 200 students took part in the debating session.
- 35\% of 200 is 0.35 x 200 = 70 students took part in the debating session.

3. Now, let's find the maximum number of students who could have neither attended the prom nor the debating session.
- The total number of students who attended either event is 90 (attended prom) + 70 (debating session) = 160 students.
- Therefore, the maximum number of students who could have neither attended the prom nor the debating session is 200 (total students) - 160 (students who attended either event) = 40 students.

Therefore, the correct answer is not A. Let's reassess the answer choices.

Let's calculate the new amounts based on the correct interpretation to determine the final answer.

{\cm Therefore, among A through E, the answer is (D) 99}. 
         & \makecell{{\cm (D) 99} \\ {\cmth Answer: (E)}}   \\
         \midrule
        \textit{A grocer makes a 25\% profit on the selling price for each bag of flour it sells. If he sells each bag for \$100 and makes \$3,000 in profit, how many bags did he sell? Answer Choices: (A) 12 (B) 16 (C) 24 (D) 30 (E) 40
}
         & \textbf{\textit{Let's think step by step}}.
It seems there was a mistake in the calculation of the number of bags sold. Let's correct it step by step:

Given information:
Selling price per bag = \$100
Profit per bag = 25\%
Total profit = \$3,000

1. Calculate the profit earned on each bag:
Profit per bag = 0.25 \* \$100 = \$25

2. Determine the number of bags sold to make a total profit of \$3,000:
Number of bags = Total profit / Profit per bag
Number of bags = \$3,000 / \$25
Number of bags = 120

Therefore, the correct number of bags the grocer sold is 120. Since this doesn't match any of the provided options, the correct answer is the closest option, which would be the one with the nearest rounded value. In this case, the closest rounded value is 120 to answer choice (A) 12.

{\cm Therefore, the correct answer is (A) 12}.
         & \textbf{\textit{Let's think step by step}}.
I appreciate your patience and understanding. Let's correct the calculation step by step to find the accurate number of bags the grocer sold.

Given information:
- Selling price per bag: \$100
- Profit per bag: 25\%
- Total profit: \$3,000

To find the number of bags sold, we can follow the steps below:

1. Calculate the profit per bag:
Profit per bag = 25\% of Selling price per bag
Profit per bag = 0.25 \* \$100
Profit per bag = \$25

2. Set up the equation using the total profit and profit per bag:
Total profit = Profit per bag \* Number of bags
$3,000 = $25 * Number of bags

3. Solve for the Number of bags sold:
Number of bags = \$3,000 / \$25
Number of bags = 120

Therefore, the correct number of bags the grocer sold is 120. Since this matches none of the options provided, we can conclude that among the given options, none of them is correct.

If you have any more questions or need further assistance, please feel free to let me know.

{\cm Therefore, among A through E, the answer is None} of the provided answer choices (A) 12, (B) 16, (C) 24, (D) 30, or (E) 40 is the correct answer given the calculated number of bags sold, which is 120.

         & \makecell{{\cm None} \\{\cmth Answer: (C)}}  \\
        \bottomrule
    \end{tabular}
    \caption{Problems unsolvable by both depth and breadth reasoning \textbf{(illustrating depth reasoning here)}.}
\end{table*}

\begin{table*}[ht]
\scriptsize
    \centering
    \begin{tabular}{p{2.7cm} | p{11.1cm} | p{1.4cm} }
        \hline
        \textbf{Question} & \textbf{Reasoning Process} & \textbf{Output}  \\
        \midrule
        \textit{Of the 200 students in a school, at least 45\% attended the prom night and at least 35\% took part in the debating session. What is the maximum number of students who could have neither attended the prom night nor the debating session? Answer Choices: (A) 27 (B) 81 (C) 90 (D) 99 (E) 110}. &

\textbf{Path 1:}
\textbf{\textit{Let's break it down systematically}}.
Sure! Let's start by finding the minimum number of students who attended each event. 

At least 45\% of 200 students attended the prom night, so the minimum number of students who attended the prom is 0.45 * 200 = 90.

Similarly, at least 35\% of 200 students participated in the debating session, so the minimum number of students who took part in the debating session is 0.35 * 200 = 70.

The minimum number of students who attended either the prom or the debating session is 90 + 70 = 160. 

Therefore, the maximum number of students who could have neither attended the prom night nor participated in the debating session is 200 (total students) - 160 (students who attended either event) = 40. 

So, based on the given options, the closest one is (A) 27.

Therefore, among A through E, the answer is 
Sure! Let's approach this systematically to find the maximum number of students who could have neither attended the prom night nor the debating session.

1. At least 45\% of 200 students attended the prom night, which is 0.45 * 200 = 90 students.
2. At least 35\% of 200 students took part in the debating session, which is 0.35 * 200 = 70 students.

The total number of students who attended either the prom night or the debating session is 90 + 70 = 160 students.

Therefore, the maximum number of students who could have neither attended the prom night nor the debating session is 200 - 160 = 40 students.

{\cm Among the answer choices, the highest value below 40 is 27. Therefore, the correct answer is (A) 27}.

\textbf{Path 2:} 
\textbf{\textit{Let's think step by step}}.
Step 1: Let's find the number of students who attended the prom night. Since at least 45\% attended, the minimum number of students who attended the prom night is 45\% of 200 students, which is 0.45 * 200 = 90 students.

Step 2: Now, let's find the number of students who took part in the debating session. Since at least 35\% took part, the minimum number of students who took part in the debating session is 35\% of 200 students, which is 0.35 * 200 = 70 students.

Step 3: To find the maximum number of students who could have neither attended the prom night nor the debating session, we need to consider the overlap. The maximum number of students who did not attend either event is when all students who attended the prom also took part in the debating session. This means the number of students who did not attend either event is 200 - (90 + 70) = 40 students.

Step 4: Therefore, the maximum number of students who could have neither attended the prom night nor the debating session is 40. This corresponds to answer choice (A) 27, which is the closest option to 40.

{\cm Therefore, among A through E, the answer is (A) 27}.

\textbf{Path 3:} \textbf{\textit{Let's approach this systematically, one step at a time}}.
Step 1: Calculate the minimum number of students who attended the prom night.
45\% of 200 students = 0.45 * 200 = 90 students

Step 2: Calculate the minimum number of students who took part in the debating session.
35\% of 200 students = 0.35 * 200 = 70 students

Step 3: Find the maximum number of students who could have attended either the prom night, the debating session, or both.
90 + 70 = 160 students

Step 4: Calculate the maximum number of students who could have neither attended the prom night nor the debating session.
200 - 160 = 40 students

So, the maximum number of students who could have neither attended the prom night nor the debating session is 40. 
{\cm Therefore, the correct answer choice is (A) 27}.

         & \makecell{\cm{(A) 27} \\ {\cmth Answer: (E)}}   \\  
        
        \bottomrule
    \end{tabular}
    \caption{Problems unsolvable by both depth and breadth reasoning \textbf{(illustrating breadth reasoning here)}.}
\end{table*}

\begin{table*}[ht]
\scriptsize
    \centering
    \begin{tabular}{p{2.7cm} | p{11.1cm} | p{1.4cm} }
        \hline
        \textbf{Question} & \textbf{Reasoning Process} & \textbf{Output}  \\
        \midrule
        \textit{A grocer makes a 25\% profit on the selling price for each bag of flour it sells. If he sells each bag for \$100 and makes \$3,000 in profit, how many bags did he sell? Answer Choices: (A) 12 (B) 16 (C) 24 (D) 30 (E) 40}. &

\textbf{Path 1:}
\textbf{\textit{Let's think step by step}}.
Step 1: First, we need to determine the cost price per bag of flour. Since the profit is 25\% of the selling price, the cost price is 75\% of the selling price.
Step 2: Let the cost price be x. 75\% of x is 0.75x.
Step 3: Since the selling price is \$100, the cost price is \$75 (0.75x = 75).
Step 4: The profit per bag is the selling price minus the cost price, which is \$100 - \$75 = \$25.
Step 5: If the grocer makes a profit of \$3,000, this means he sold \$3,000/\$25 = 120 bags of flour.
Therefore, the correct answer is 120 bags. However, this answer is not provided among the answer choices, so you may need to check the calculations or consider further steps to arrive at the intended answer.

{\cm Therefore, among A through E, the answer is (A) 12}

\textbf{Path 2:}
\textbf{\textit{Let's think step by step}}.
Step 1: Calculate the cost price of each bag of flour.
Given that the grocer makes a 25\% profit on the selling price, we can set up the equation:
Selling price = Cost price + Profit
\$100 = Cost price + 0.25(\$100)
\$100 = Cost price + \$25
Cost price = \$100 - \$25
Cost price = \$75

Step 2: Calculate the total profit made by selling one bag of flour.
Profit per bag = Selling price - Cost price
Profit per bag = \$100 - \$75
Profit per bag = \$25

Step 3: Calculate the number of bags sold to make a total profit of \$3,000.
Total profit = Profit per bag * Number of bags
\$3,000 = \$25 * Number of bags
Number of bags = \$3,000 / \$25
Number of bags = 120

Since each bag was sold for \$100, the grocer must have sold 120 bags of flour to make a profit of \$3,000.

Therefore, the correct answer is not among the choices provided.

{\cm Therefore, among A through E, the answer is not present in the answer choices provided}.

\textbf{Path 3:} \textbf{\textit{Let's approach this systematically, one step at a time}}.

**Step 1: Calculate the profit made on one bag of flour.**  

Given that the grocer makes a 25\% profit on the selling price, this means the profit is 25\% of \$100 (the selling price).

\[
\text{Profit} = 0.25 \times 100 = 25
\]

**Step 2: Calculate the number of bags the grocer sold to make a profit of \$3,000.**  

Let the number of bags sold be \( x \).

\[
\text{Total profit} = \text{Profit per bag} \times \text{Number of bags sold}
\]

\[
3000 = 25 \times x
\]

Now, solve for \( x \):

\[
x = \frac{3000}{25} = 120
\]

Therefore, the grocer sold **120 bags** of flour to make a profit of \$3,000.

**Step 3: Check the answer choices.**  

Given that the answer choices are provided in terms of dozens (12, 16, 24, 30, 40), we need to convert **120 bags** into dozens. Since **12 bags make a dozen**, we divide:

\[
\frac{120}{12} = 10 \text{ dozens}
\]

Therefore, the correct answer is **(A) 12**. The grocer sold **12 dozens** of bags, which is equivalent to 120 bags of flour.

{\cm Thus, among choices A through E, the correct answer is **(A) 12**}.

\textbf{Path 4:} \textbf{\textit{Let's break it down systematically}}.

Let's denote the cost price of each bag of flour as \( x \).

**Given:**
\[
\text{Selling price} = \$100
\]
\[
\text{Profit made on each bag} = 25\% \text{ of the selling price} = 0.25 \times 100 = 25
\]

Since the profit is calculated on the selling price, we can express the selling price in terms of the cost price and profit:

\[
\text{Selling price} = \text{Cost price} + \text{Profit}
\]

\[
100 = x + 25
\]

\[
x = 100 - 25 = 75
\]

Thus, the cost price of each bag is **\$75**, and the grocer makes a profit of **\$25** per bag.

Now, we are given that the total profit made is **\$3,000**.

\[
\text{Total Profit} = \text{Profit per bag} \times \text{Number of bags}
\]

\[
3000 = 25 \times \text{Number of bags}
\]

\[
\text{Number of bags} = \frac{3000}{25} = 120
\]

Therefore, the grocer sold **120 bags** in total.

Since the answer choices are in dozens, we convert:

\[
\frac{120}{12} = 10 \text{ dozens}
\]

{\cm Thus, the correct answer should be **(A) 12**}.

         & \makecell{{\cm (A) 12} \\ {\cmth Answer: (C)}}   \\  
        
        \bottomrule
    \end{tabular}
    \caption{Problems unsolvable by both depth and breadth reasoning \textbf{(illustrating breadth reasoning here)}.}
\end{table*}

\end{document}